\documentclass[preprint,12pt]{elsarticle}

\usepackage{amssymb}
\usepackage{amsmath}
\usepackage{tikz}
\usepackage{forest}
\usetikzlibrary{arrows.meta, positioning, shapes.multipart, trees}

\begin{document}

\begin{frontmatter}

\title{Natural Language Processing for Analyzing Electronic Health Records and Clinical Notes in Cancer Research: A Review}

\author[label1]{Muhammad Bilal\corref{cor1}}
\author[label2]{Ameer Hamza} 
\author[label3]{Nadia Malik} 

\affiliation[label1]{organization={Department of Pharmaceutical Outcomes and Policy, University of Florida},
            city={Gainesville},
            postcode={32611}, 
            state={Florida},
            country={USA}}
\affiliation[label2]{organization={Department of Computer Science, Faculty of Computing and IT, University of Sargodha},
            city={Sargodha},
            postcode={40100}, 
            state={Punjab},
            country={Pakistan}}
\affiliation[label3]{organization={Department of Software Engineering, Faculty of Computing and IT, University of Sargodha},
            city={Sargodha},
            postcode={40100}, 
            state={Punjab},
            country={Pakistan}}
            
\cortext[cor1]{muhammad.bilal@ufl.edu; mbilal.csit@gmail.com (Muhammad Bilal)}

\begin{abstract}
\textbf{Objective:} This review aims to analyze the application of natural language processing~(NLP) techniques in cancer research using electronic health records~(EHRs) and clinical notes. This review addresses gaps in the existing literature by providing a broader perspective than previous studies focused on specific cancer types or applications.
\\
\textbf{Methods:} A comprehensive literature search was conducted using the Scopus database, identifying 94 relevant studies published between 2019 and 2024. Data extraction included study characteristics, cancer types, NLP methodologies, dataset information, performance metrics, challenges, and future directions. Studies were categorized based on cancer types and NLP applications.
\\
\textbf{Results:} The results showed a growing trend in NLP applications for cancer research, with breast, lung, and colorectal cancers being the most studied. Information extraction and text classification emerged as predominant NLP tasks. A shift from rule-based to advanced machine learning techniques, particularly transformer-based models, was observed. The Dataset sizes used in existing studies varied widely. Key challenges included the limited generalizability of proposed solutions and the need for improved integration into clinical workflows.
\\
\textbf{Conclusion:} NLP techniques show significant potential in analyzing EHRs and clinical notes for cancer research. However, future work should focus on improving model generalizability, enhancing robustness in handling complex clinical language, and expanding applications to understudied cancer types. Integration of NLP tools into clinical practice and addressing ethical considerations remain crucial for utilizing the full potential of NLP in enhancing cancer diagnosis, treatment, and patient outcomes.

\end{abstract}


\end{frontmatter}

\section{Introduction}

Cancer remains one of the most significant global health challenges, with recent projections indicating 1,958,310 new cancer cases and 611,720 cancer deaths in the United States for 2024 \cite{siegel2024cancer}. Despite significant advancements in cancer research and treatment, there is an urgent need for innovative approaches to improve patient outcomes and address persistent disparities in cancer care \cite{schepis2023state,lu2023landscape}. Traditionally, cancer research has heavily relied on structured data and imaging techniques, such as histopathology and radiology \cite{shmatko2022artificial,veta2014breast}. Advanced image processing and computer vision techniques have been applied to analyze medical imaging data to assist in early detection and accurate diagnosis of various cancers \cite{bi2019artificial,elemento2021artificial}. While these techniques have produced useful results, they frequently fail to capture the rich, complex information contained in unstructured clinical text. Electronic health records~(EHRs) and clinical notes contain extensive information about patient's medical histories, diagnoses, treatments, and outcomes, providing a unique opportunity to gain a better understanding of cancer progression, treatment effectiveness, and patient experiences \cite{sitapati2017integrated,jensen2012mining}. 

In recent years, the application of natural language processing~(NLP) to analyze EHRs and clinical notes has emerged as a promising field for advancing cancer research \cite{kreimeyer2017natural}. NLP techniques can automatically analyze large volumes of clinical text, identify relevant information, and generate structured data for further analysis \cite{tayefi2021challenges}. This capability has the potential to revolutionize cancer research by enabling the extraction of valuable insights from enormous amounts of previously unexplored clinical data \cite{datta2019frame}. Moreover, by automating the analysis of unstructured clinical data, NLP can substantially reduce the costs and time associated with manual data review. This efficiency can expedite the review of extensive medical records and potentially identify early warning signs leading to earlier and more accurate diagnoses. Furthermore, by improving the accuracy of data extraction and analysis, NLP can enhance clinical decision-making and improve patient health outcomes.

In recent years, the application of NLP to analyze EHRs and clinical notes in cancer research has gained significant attention. However, compared to reviews on image processing and computer vision techniques for cancer research, very few reviews exist on NLP applications in cancer research. While recent studies have attempted to address this gap by exploring various aspects of NLP in cancer research they have primarily concentrated on specific areas or cancer types. For instance, \citet{datta2019frame} provided a frame semantic overview of NLP-based information extraction for cancer-related EHR notes, while \citet{aggarwal2024advancements} examined advancements and challenges in NLP for oral cancer research. \citet{wang2022assessment} assessed EHRs for cancer research and patient care through a scoping review, and \citet{gholipour2023extracting} systematically reviewed studies that applied NLP methods to identify cancer concepts from clinical notes. Additionally, \citet{bitterman2021clinical} presented a review of clinical NLP for radiation oncology, and \citet{sangariyavanich2023systematic} conducted a systematic review of NLP for recurrent cancer detection from electronic medical records.

The limitations attached to existing reviews highlight the need for a more comprehensive review that can provide valuable insights into various cancer types, NLP applications, techniques, datasets, and challenges. This review aims to address this gap by analyzing NLP applications in cancer research using EHRs and clinical notes, classifying studies based on different cancer types, discussing NLP techniques and applications, examining available datasets, and exploring open research challenges and future directions. This will give researchers and clinicians a broader understanding of the current state and potential future developments.

The remainder of this paper is structured as follows: Section~2 details the methodology used in conducting this review. Section~3 provides a taxonomy and categorizes the studies by cancer type and research objective. Section~4 discusses the overall distribution of studies across different categories and journals highlighting trends in the number of publications, and identifying top-cited papers. Section~5 discusses the various NLP techniques and methodologies identified in the literature. Section~6 examines NLP-based datasets in cancer research, discussing their characteristics, availability, and potential for future research. Section~7 outlines open research challenges and future directions for NLP in cancer research. Lastly, Section~8 concludes this review with a summary of key findings and implications for future research.

\section{Materials and Methods}

This study conducted a systematic literature review to investigate the use of NLP techniques in analyzing EHRs for cancer research. The review focused on peer-reviewed articles published since 2019. The literature search was conducted using the ``Scopus'' database. The search query was carefully created to capture relevant studies addressing the intersection of EHRs, cancer, and NLP. The search query included key terms such as ``Electronic Health Records''. ``EHRs'', ``Clinical Notes'', ``Cancer'', ``Natural Language Processing'', and ``NLP''. The query was structured to focus on these concepts within the title, abstract, and keywords of articles. The selection of studies for this review was guided by a set of predefined inclusion and exclusion criteria. To be included, studies had to be published in peer-reviewed journals since 2019, focusing specifically on the application of NLP techniques to EHRs or clinical notes in the context of cancer. Only articles written in English were considered. The review excluded conference papers, book chapters, reviews, and editorials to maintain a focus on primary research articles. Studies that did not directly address the use of NLP on EHR data for cancer research were also excluded. Additionally, articles published outside the specified time frame were not considered. These criteria were strictly applied throughout the screening process to ensure the inclusion of articles that specifically addressed the application of NLP techniques to EHRs or clinical notes in the context of cancer, published in reputable academic journals within the specified time frame.

\begin{figure}[!b]
    \centering
    \includegraphics[width=.80\textwidth]{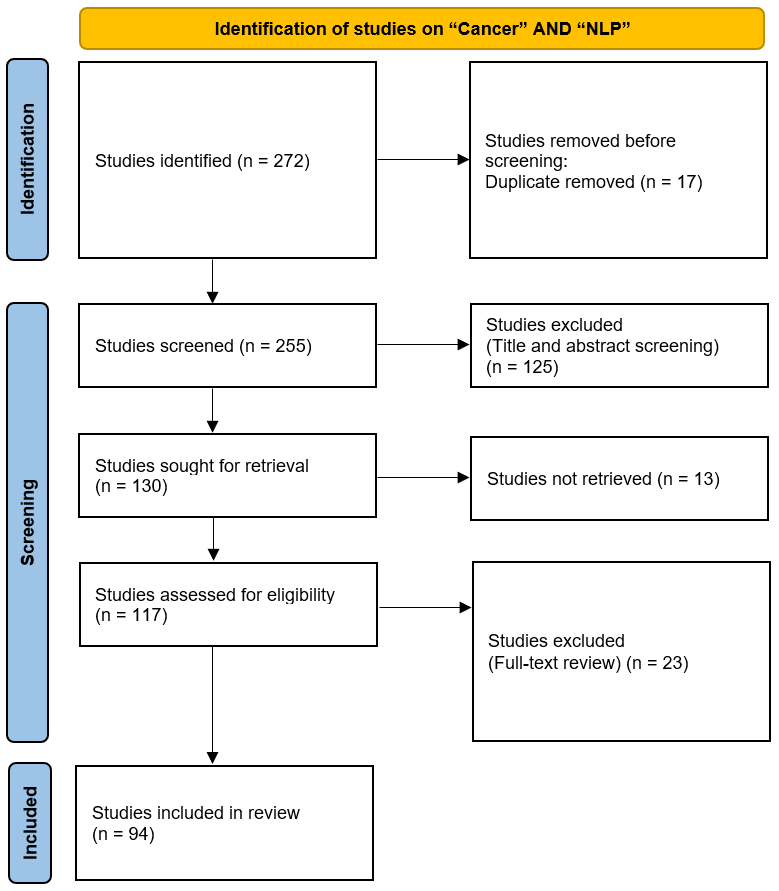}
    \caption{Prisma flow diagram for selection of studies.}
    \label{fig:Prisma_flow}
\end{figure}

The flow diagram in Figure~\ref{fig:Prisma_flow} illustrates the selection process, clearly indicating the number of studies at each stage of screening and the reasons for exclusions. Initially, the Scopus database search identified 272 potentially relevant studies. After removing duplicate studies 255 articles underwent a preliminary screening based on titles and abstracts, resulting in 130 studies being considered for further review. Subsequently, 130 articles were attempted to be retrieved for full-text review, and 117 studies were successfully received for full-text evaluation. Finally, 94 studies were selected to be included in this review. This rigorous selection process ensured that only the most relevant and high-quality studies addressing the application of NLP to EHRs in cancer care were included in the final analysis. 

This review was guided by key research questions focusing on the categorization of studies by cancer types, distribution of research across cancer categories, predominant NLP applications and techniques, NLP-based datasets, and current research challenges and future directions. For each selected study, we extracted a comprehensive set of data including study identifiers, characteristics, cancer-specific information, NLP-related details, dataset information, outcomes and performance metrics, reported challenges and limitations, and suggested future directions. This systematic extraction process enabled us to develop a detailed taxonomy based on cancer types and facilitated in-depth analysis of NLP applications in cancer research using EHRs and clinical notes. This provided the foundation for our discussion on study distribution, NLP techniques, datasets, and open research challenges.

\section{Taxonomy of NLP in Cancer Research}
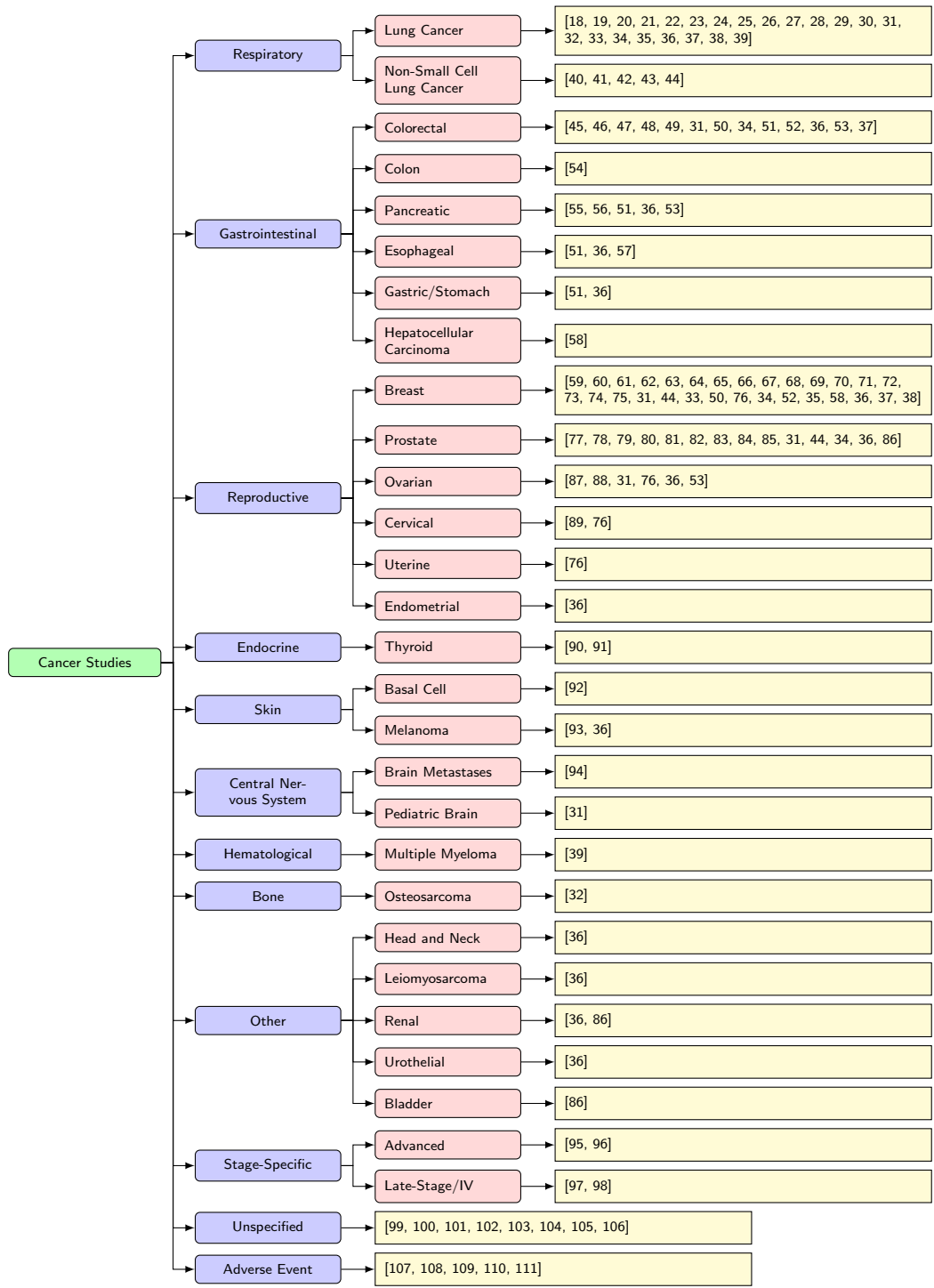
\begin{figure}
\centering
    
\tikzset{
    basic/.style  = {draw, text width=2cm, align=center, font=\sffamily\tiny, rectangle},
    root/.style   = {basic, rounded corners=2pt, thin, align=center, fill=green!30},
    onode/.style = {basic, thin, rounded corners=2pt, align=center, fill=green!60, text width=1.9cm,},
    tnode/.style = {basic, thin, align=left,rounded corners=2pt, fill=pink!60, text width=1.9cm, align=left},
    snode/.style = {basic, thin, align=left, fill=yellow!20, text width=5.35cm, align=left},
    xnode/.style = {basic, thin, rounded corners=2pt, align=center, fill=blue!20,text width=1.9cm,},
    edge from parent/.style={draw=black, edge from parent fork right, font=\tiny}
}

\begin{forest} for tree={
    grow=east,
    reversed,
    growth parent anchor=west,
    growth parent anchor=west,
    parent anchor=east,
    child anchor=west,
    edge path={\noexpand\path[\forestoption{edge},->, >={latex}] 
         (!u.parent anchor) -- +(5pt,0pt) |-  (.child anchor) 
         \forestoption{edge label};},
    l sep=5mm,
    s sep=1.2mm,
    anchor=west,
    font=\tiny
}
[Cancer Studies, root
    [Respiratory, xnode
        [Lung Cancer, tnode
            [\cite{hong2020annotation, ruckdeschel2023unstructured, hu2024zero, solarte2021integrating, gauthier2022automating, liu2023leveraging, yang2024extracting, ebrahimi2024identification, wang2019natural, yu2021study, zigman2022timely, yu2022assessing, benedum2023replication, hochheiser2023deepphe, huang2024critical, morin2021artificial, kehl2021clinical, bhatt2023use, kehl2021artificial, yu2024identifying, araki2023real, wang2019achievability}, snode]]
        [Non-Small Cell Lung Cancer, tnode
            [\cite{paul2022investigation, kersloot2019automated, adamson2023approach, yusuf2024text, seesaghur2023assessment}, snode]]
    ]
    [Gastrointestinal, xnode
        [Colorectal, tnode
            [\cite{li2023interpretable, laique2021application, agaronnik2020challenges, tamm2022establishing, karwa2020development, hochheiser2023deepphe, luo2021computational, kehl2021clinical, do2021patterns, luo2021analyzing, kehl2021artificial, shi2022identifying, yu2024identifying}, snode]]
        [Colon, tnode
            [\cite{ryu2020transformation}, snode]]
        [Pancreatic, tnode
            [\cite{sarwal2024identification, harrison2023successful, do2021patterns, kehl2021artificial, shi2022identifying}, snode]]
        [Esophageal, tnode
            [\cite{do2021patterns, kehl2021artificial, chen2023natural}, snode]]
        [Gastric/Stomach, tnode
            [\cite{do2021patterns, kehl2021artificial}, snode]]
        [Hepatocellular\\Carcinoma, tnode
            [\cite{karimi2021development}, snode]]
    ]
    [Reproductive, xnode
        [Breast, tnode
            [\cite{zhou2023cross, zhou2022cancerbert, garcia2024gpt, ribelles2021machine, sanyal2021weakly, thompson2019relevant, zeng2019identifying, jin2021research, chen2022automated, brizzi2020natural, alkaitis2021automated, solarte2023transformers, zelina2023extraction, trivedi2019large, levine2019learning, meystre2019automatic, santus2019neural, hochheiser2023deepphe, seesaghur2023assessment, morin2021artificial, luo2021computational, zhao2021generating, kehl2021clinical, luo2021analyzing, bhatt2023use, karimi2021development, kehl2021artificial, yu2024identifying, araki2023real}, snode]]
        [Prostate, tnode
            [\cite{yang2022identification, alba2021ascertainment, hernandez2020leveraging, bozkurt2022expanding, bozkurt2020phenotyping, bozkurt2019possible, coquet2019comparison, zhu2019automatically, banerjee2019weakly, hochheiser2023deepphe, seesaghur2023assessment, kehl2021clinical, kehl2021artificial, huang2023natural}, snode]]
        [Ovarian, tnode
            [\cite{mcgowan2024can, laios2023roberta, hochheiser2023deepphe, zhao2021generating, kehl2021artificial, shi2022identifying}, snode]]
        [Cervical, tnode
            [\cite{macchia2022multidisciplinary, zhao2021generating}, snode]]
        [Uterine, tnode
            [\cite{zhao2021generating}, snode]]
        [Endometrial, tnode
            [\cite{kehl2021artificial}, snode]]
    ]
    [Endocrine, xnode
        [Thyroid, tnode
            [\cite{yoo2022transforming, pathak2023extracting}, snode]]
    ]
    [Skin, xnode
        [Basal Cell, tnode
            [\cite{ali2022development}, snode]]
        [Melanoma, tnode
            [\cite{malke2019enhancing, kehl2021artificial}, snode]]
    ]
    [Central Nervous System, xnode
        [Brain Metastases, tnode
            [\cite{senders2019natural}, snode]]
        [Pediatric Brain, tnode
            [\cite{hochheiser2023deepphe}, snode]]
    ]
    [Hematological, xnode
        [Multiple Myeloma, tnode
            [\cite{wang2019achievability}, snode]]
    ]
    [Bone, xnode
        [Osteosarcoma, tnode
            [\cite{huang2024critical}, snode]]
    ]
    [Other, xnode
        [Head and Neck, tnode
            [\cite{kehl2021artificial}, snode]]
        [Leiomyosarcoma, tnode
            [\cite{kehl2021artificial}, snode]]
        [Renal, tnode
            [\cite{kehl2021artificial, huang2023natural}, snode]]
        [Urothelial, tnode
            [\cite{kehl2021artificial}, snode]]
        [Bladder, tnode
            [\cite{huang2023natural}, snode]]
    ]
    [Stage-Specific, xnode
        [Advanced, tnode
            [\cite{lindvall2022natural, lindvall2019natural}, snode]]
        [Late-Stage/IV, tnode
            [\cite{ernecoff2019electronic, dimartino2022identification}, snode]]
    ]
    [Unspecified, xnode
        [\cite{laurent2023automatic, lin2023machine, naseri2021development, ahmad2023bir, de2022class, cohen2023natural, koleck2021characterizing, guan2019natural}, snode]]
    [Adverse Event, xnode
        [\cite{mashima2022using, li2022developing, hong2020natural, munoz2023development, zitu2023generalizability}, snode]]
]
\end{forest}
    \caption{Taxonomy of Cancer Types and respective studies.}
    \label{fig:cancer_taxonomy}
\end{figure}

This section presents a taxonomy of NLP applications in cancer research, categorized based on a thorough review of research objectives and cancer types. Figure~\ref{fig:cancer_taxonomy} provides a visual representation of the overall taxonomy. Moreover, studies that cover multiple cancer types are discussed separately in this section under ``Multiple Cancer Types''.
\subsection{Respiratory Cancers}

\subsubsection{Lung Cancer}
Several studies have focused on extracting specific types of information from clinical notes for lung cancer research. \citet{hong2020annotation} developed a rule-based Named Entity Recognizer to extract age and temporal events from clinical histories, achieving F1-scores of 86\%. \citet{ruckdeschel2023unstructured,liu2023leveraging} and \citet{ebrahimi2024identification} have developed methods to extract smoking status and history from clinical notes, resulting in high F-scores and improved patient identification for screening. Extraction of lung cancer diagnosis information has also been explored, with \cite{solarte2021integrating} achieving F1-scores of 90\% for named entity recognition and 89\% for relating diagnoses to dates in Spanish clinical notes. Studies by \citet{gauthier2022automating} and \citet{wang2019natural} investigated the extraction of various clinical details, demonstrating high accuracy and recall rates for demographic data and cancer-related information. Radiological information extraction was examined by \citet{yang2024extracting}, comparing transformer models for pulmonary nodule information extraction, with RoBERTa-mimic achieving the best F1-score of 0.9279. Recent research has explored large language models for information extraction, with \citet{hu2024zero} investigating ChatGPT's performance on lung cancer radiology reports. Social and behavioral determinants of health (SBDoH) extraction has been studied by \citet{yu2021study} and \citet{yu2022assessing}, using bidirectional encoder representations from transformers~(BERT) based models and achieved high F1-scores. \citet{zigman2022timely} analyzed EHRs to investigate the timeliness of lung cancer diagnosis, revealing a median interval of 570 days from first symptoms to diagnosis. Finally, \citet{benedum2023replication} evaluated the quality of oncology data extracted using machine learning and NLP techniques, finding less than 8\% difference compared to expert manual abstraction.

\subsubsection{Non-Small Cell Lung Cancer (NSCLC)}
A number of studies have also demonstrated the effectiveness of automated systems in extracting and analyzing NSCLC-related information from clinical data. An automated medical reports processing system developed by \citet{kersloot2019automated}, showed high performance in detecting lung cancer and NSCLC concepts, with F1-scores ranging from 0.828 to 0.947 and 0.862 to 0.933, respectively. Similarly, \citet{yusuf2024text} created a text analysis framework for extracting epidermal growth factor receptor~(EGFR) mutation information from unstructured clinical data, achieving 97.5\% accuracy on a sample of 362 tests. Expanding on these efforts, \citet{adamson2023approach} applied NLP and machine learning methods to a nationwide EHR-derived database of over 300,000 cancer patients. This study developed models to extract various cancer-related variables with high performance, including 0.96 sensitivity and 0.92 positive predictive value for extracting non-squamous histology in NSCLC patients. Addressing data protection concerns, \citet{paul2022investigation} built a clinical de-identification model using conditional random fields (CRF) for named entity recognition on EHRs. Analyzing 1,500 pathology reports from 421 advanced NSCLC patients, they found that n-gram, prefix-suffix, word embedding, and word shape features performed best, with the model achieving F1-scores ranging from 0.65 to 0.95 for most protected health information categories, and notably observed that performance was saturated with around 200 training samples.

\subsection{Gastrointestinal Cancers}
\subsubsection{Colorectal Cancer}
Colorectal cancer remains a significant health concern, prompting researchers to explore advanced approaches for improving diagnosis, treatment, and follow-up care. Research at Cleveland Clinic and the University of Minnesota \cite{laique2021application} evaluated a hybrid optical character recognition and NLP approach for extracting quality metric data from colonoscopy and pathology reports, demonstrating high accuracy in detecting various colonoscopy-related factors. Similarly, a study involving three national health service~(NHS) Trusts in the UK \cite{tamm2022establishing} developed a research database for colorectal cancer using routinely collected health data, implementing NLP to extract relevant information from imaging and histopathology reports. Additionally, researchers have explored the application of NLP in clinical decision support and predictive modeling. A study at Cleveland Clinic \cite{karwa2020development} developed and validated an automated clinical decision support algorithm for follow-up colonoscopy recommendations, achieving 100\% accuracy in data extraction and 99\% guideline concordance.

\citet{li2023interpretable} developed an interpretable fusion model to predict liver metastases in postoperative colorectal cancer patients, achieving high predictive performance by integrating both free-text medical records and structured laboratory data. A study also attempted to assess the utility of NLP in identifying individuals with chronic mobility disability from electronic health records. \citet{agaronnik2020challenges} analyzed 303,182 clinical notes from 14,877 colorectal cancer patients. Although NLP screening identified notes containing wheelchair-associated keywords, only a small portion contained clear documentation of wheelchair use reason and duration. This research highlighted the need for manual chart review to confirm chronic mobility disability and emphasized inadequate disability documentation in many clinical notes.

\subsubsection{Colon Cancer}
\citet{ryu2020transformation} focused on transforming colon cancer pathology reports into the observational medical outcomes partnership~(OMOP) common data model~(CDM) format. The researchers utilized NLP techniques to extract and standardize clinical text entities from pathology reports, including surgical specimens, immunohistochemical studies, and molecular studies. The dataset comprised 18,090 reports from Seoul National University Bundang Hospital. The extracted information was mapped to standard terminologies and inserted into CDM tables. The study achieved 100\% recognition accuracy for key attributes from surgical specimen pathology and molecular study reports, successfully creating a standardized database for colon cancer research. This transformation enables the integration of pathology data with other clinical and omics data, facilitating more comprehensive cancer research using the OMOP CDM framework.

\subsubsection{Pancreatic Cancer}
\citet{sarwal2024identification} aimed to use NLP to automatically extract pancreatic ductal adenocarcinoma (PDAC) risk factors from unstructured clinical notes in EHRs. The researchers developed rule-based NLP algorithms and tested them on a dataset of 2091 clinical notes and an additional cohort of 1138 patients. The algorithms showed high sensitivity for identifying PDAC risk factors, with particularly strong results for family history detection. For the family history of PDAC, the algorithm achieved a recall of 0.933, precision of 0.790, and F1-score of 0.856. For germline genetic mutations, the recall was 0.851, but precision (0.350) and F1-score (0.496) were lower due to false positives from tissue mutation misidentification. The study suggests that NLP can be a valuable tool for automating the identification of high-risk individuals for pancreatic cancer screening, though further validation in larger primary-care populations is recommended.

\subsection{Reproductive System Cancers}
\subsubsection{Breast Cancer}
The use of NLP to extract breast cancer phenotypes from EHRs has been investigated in many studies. \citet{zhou2023cross} evaluated the generalizability of different NLP models, comparing conditional random field~(CRF), bidirectional long short-term memory CRF (BiLSTM-CRF), and CancerBERT models. Using a large dataset of 72,020 breast cancer patient records, the study found that CancerBERT models demonstrated better performance and generalizability. \citet{zhou2022cancerbert} proposed and evaluated CancerBERT, achieving high macro F1-scores for exact and lenient matches. In the context of non-English EHRs, researchers have explored NLP applications in various languages. Studies on Spanish \cite{garcia2024gpt, solarte2023transformers} and Czech \cite{zelina2023extraction} clinical notes have shown promising results using different transformer-based models and transfer learning techniques.

Several studies have focused on predicting breast cancer outcomes and treatment responses using NLP and machine learning techniques. \citet{ribelles2021machine} investigated predictive models for early and late progression to first-line treatment in hormone receptor-positive (HR+)/HER2-negative advanced breast cancer patients, with NLP-based approaches slightly outperforming manual feature extraction. Another study by \citet{sanyal2021weakly} developed a weakly supervised deep learning model for predicting breast cancer distant recurrence, achieving high area under the receiver operating characteristic curve~(AUROC), sensitivity, and specificity. Research has also been conducted on extracting specific information from pathology reports and clinical notes. A novel NLP technique called relevant word order vectorization (RWOV) \cite{thompson2019relevant} was introduced for analyzing EHRs, achieving high F1-scores when classifying hormone receptor status in breast cancer patients. The application of NLP in clinical trial recruitment has been explored by \citet{meystre2019automatic}, with machine learning-based NLP applications demonstrating high recall and precision in extracting eligibility criteria.

The creation of extensive databases and systems for breast cancer research has been examined by several studies. A study by \citet{jin2021research} described the creation of a breast cancer-specific database system, incorporating data standards, fusion, governance, and security measures. The cross-institutional generalizability of NLP algorithms has also been investigated by \citet{santus2019neural}, demonstrating good generalizability across institutions. Some studies have also explored the use of NLP in specific clinical contexts, such as examining the impact of palliative care consultations on end-of-life care measures \cite{brizzi2020natural} and extracting treatment discontinuation rationale from EHRs \cite{alkaitis2021automated}. Moreover, semi-automated methods for labeling clinical records have been explored, comparing traditional NLP techniques with IBM Watson's Natural Language Classifier \cite{trivedi2019large} and developing learning health system prototypes \cite{levine2019learning}.

\subsubsection{Prostate Cancer}
Existing studies have utilized NLP to assess various aspects of prostate cancer patient care and outcomes and developed NLP models to identify patients with metastatic prostate cancer (mPCa) from clinical records. \citet{yang2022identification} developed an NLP model using a large dataset of radiology reports from the Department of Veterans Affairs, demonstrating high accuracy in identifying mPCa patients. Another study by \citet{alba2021ascertainment} validated an NLP algorithm to identify mPCa patients in the Veterans Affairs EHR, achieving high specificity and sensitivity. NLP techniques have also been applied to extract cancer staging information from clinical notes. Furthermore, \citet{bozkurt2022expanding} compared rule-based and machine learning approaches for extracting TNM stages from clinical notes of prostate cancer patients, finding that different approaches performed better for different stages. \citet{hernandez2020leveraging} used NLP and machine learning techniques to analyze EHR data, demonstrating high precision and recall rates for identifying digital rectal examinations and classifying anesthesia types. Similarly, \citet{bozkurt2019possible} investigated the feasibility of automatically assessing pretreatment digital rectal examination reports using NLP.

NLP techniques have been applied to evaluate guideline adherence and treatment decisions. \citet{coquet2019comparison} developed an automated pipeline to assess bone scan utilization in prostate cancer patients. The study compared rule-based and convolutional neural networks~(CNN) based NLP approaches for classifying patients into risk groups and evaluating adherence to clinical guidelines. Researchers have also explored the use of NLP to extract information on patient-centered outcomes and social determinants of health. A study by \citet{bozkurt2020phenotyping} developed an NLP approach to phenotype urinary incontinence severity in prostate cancer patients, while another study by \citet{zhu2019automatically} used NLP to identify social isolation in prostate cancer patients. Both studies demonstrated high accuracy in their respective tasks. Moreover, \citet{banerjee2019weakly} developed a weakly supervised NLP approach to assess patient outcomes in prostate cancer treatment from clinical notes, achieving high performance in classifying sentences regarding urinary incontinence and bowel dysfunction.

\subsubsection{Ovarian Cancer}
A study by \citet{mcgowan2024can} conducted at a UK cancer center evaluated the use of NLP for enhancing the efficiency of gynaecological oncology audits. The researchers utilized Google Bard to develop algorithms for analyzing the EHRs of ovarian cancer patients. Comparing NLP-generated code against previous manual audits, they found that the NLP approach significantly increased efficiency by enabling faster analysis of a larger dataset of 600 cases compared to the manual audit of 135 cases. In another study \cite{laios2023roberta}, researchers investigated the use of NLP to predict surgical outcomes in advanced epithelial ovarian cancer~(EOC) patients undergoing cytoreductive surgery. They analyzed operative notes from 555 EOC cases performed between 2014 and 2019 using a RoBERTa-based classifier. The NLP model achieved high-performance metrics, including an AUROC of 0.86 and an accuracy of 0.81, outperforming models using only discrete clinical features.

\subsubsection{Cervical Cancer}
\citet{macchia2022multidisciplinary} developed a prototype of a ``Multidisciplinary Tumor Board Smart Virtual Assistant'' for locally advanced cervical cancer~(LACC). The system aimed to automate clinical stage classification from free-text diagnostic reports, identify inconsistencies, support education, and integrate data-driven decision-making. Using EHRs of 96 LACC patients treated between 2015 and 2018, the system used NLP to analyze and categorize patient data from Magnetic resonance imaging~(MRI), gynecologic examination, and positron emission tomography-computed tomography~(PET-CT) reports. The virtual assistant successfully classified all patients in the training set and developed a predictive model for patient achieving 94\% accuracy. The study demonstrates the potential for creating a smart virtual assistant to aid multidisciplinary tumor boards in managing large volumes of patient data and improving diagnostic accuracy for LACC cases.

\subsection{Endocrine Cancers}
A study by \citet{yoo2022transforming} aimed to transform unstructured thyroid cancer diagnosis and staging information into the standardized OMOP CDM format. Using rule-based NLP on 52,133 surgical pathology reports and 56,239 iodine whole-body scan reports from three medical institutions, the researchers successfully converted the extracted data to OMOP CDM v6.0. Their findings revealed that papillary carcinoma was the most prevalent thyroid cancer type having 95.3-98.8\%, with stage I accounting for 55-64\% of cases and stage III representing 24-26\%. The study achieved high precision and recall in the extraction and conversion process. Another study by \citet{pathak2023extracting}, focused on extracting thyroid nodule characteristics from ultrasound reports using transformer-based NLP methods. They compared five transformer models on a corpus of 490 reports from a dataset of 184,560 clinical notes. The GatorTron model, trained on over 90 billion words, demonstrated the best performance with strict and lenient F1-scores of 0.8851 and 0.9495 for characteristic extraction, and an F1-score of 0.9321 for linking characteristics to nodules. This research represents a pioneering effort in applying transformer-based NLP models to extract multiple thyroid nodule characteristics from ultrasound reports.

\subsection{Skin Cancers}
A couple of studies have demonstrated the potential of NLP for Skin cancer research. A study by \citet{ali2022development}, developed and validated an automated information extraction system for basal cell carcinoma (BCC) histopathology reports using the general architecture for text engineering~(GATE) framework. The system achieved high performance across various metrics, including a mean F1-score of 84.5\%, when compared to expert human annotation. Moreover, a study by \citet{malke2019enhancing} focused on melanoma research, developing an efficient method to extract primary pathology features from pathology reports. This system was applied to a large dataset of 23,039 patients, generating over 368,000 individual melanoma data points. The algorithm achieved a 90.4\% exact or synonymous match rate when compared to manually curated data, with a low error rate of 3.7\%.

\subsection{Central Nervous System (CNS) Cancers}
\citet{senders2019natural} attempted to compare various NLP techniques for automatically quantifying brain metastases from free-text radiology reports. The researchers analyzed 1,479 MRI reports of patients diagnosed with brain metastases from two tertiary care centers. They developed and evaluated multiple bag-of-words and sequence-based NLP models to classify reports as indicating single or multiple metastases. The least absolute shrinkage and selection operator~(LASSO) regression model performed best overall on the test set, achieving an AUROC of 0.92 and an accuracy of 83\%. Among sequence-based models, a 1D CNN showed the highest performance. The study provides a framework for developing machine learning-based NLP models for medical text analysis and demonstrates their potential for automated extraction of clinically relevant information from radiology reports.

\subsection{Stage-Specific Categories}
\subsubsection{Advanced Cancer}
Existing studies have highlighted the potential of NLP in improving data extraction and analysis in advanced cancer care for enhancing patient management and quality of care. \citet{lindvall2022natural} evaluated NLP for identifying advance care planning documentation in clinical notes of advanced cancer patients, achieving high accuracy with F1-scores ranging from 0.84 to 0.97 across various domains. This method proved significantly more time-efficient than manual chart review. Additionally, \citet{lindvall2019natural} aimed to use NLP to identify cancer patients receiving palliative gastrostomy and assess end-of-life quality indicators. The NLP approach demonstrated a high sensitivity of 95.8\% and specificity of 97.4\% in identifying patients with palliative indications, while also accurately assessing end-of-life process measures.

\subsubsection{Late-Stage / Stage IV Solid Tumor}
Several recent research has focused on leveraging NLP to improve patient identification and symptom management in late-stage or stage IV solid tumors. A study by \citet{ernecoff2019electronic} developed and validated EHR phenotypes to identify patients with late-stage diseases, including stage 4 solid-tumor cancer. Their approach, utilizing international classification of diseases~(ICD) codes, NLP, and laboratory values, achieved a high sensitivity of 99.5\% and a positive predictive value of 68.6\% for stage 4 cancer, demonstrating the potential for efficient patient identification and equitable distribution of palliative care. Based on computational techniques in oncology, \citet{dimartino2022identification} investigated the application of NLP to identify uncontrolled symptoms in hospitalized cancer patients with advanced disease. Using a Random Forest algorithm to analyze clinical notes, they developed models to predict uncontrolled pain, nausea/vomiting, and dyspnea, with varying levels of accuracy and sensitivity. The pain model performed better than the other models, achieving 61\% accuracy and 69\% sensitivity.

\subsection{Multiple Cancer Types}
A significant number of studies have explored the application of NLP techniques to analyze and extract valuable information from clinical data across various cancer types. \citet{hochheiser2023deepphe} developed DeepPhe-CR, an NLP system designed to help cancer registrars abstract case details from patient records, achieving high F1-scores for extracting key variables across multiple cancer types. Similarly, \citet{harrison2023successful} developed an NLP algorithm for interpreting pancreatic pathology reports, demonstrating high accuracy and potential for enhancing EHR coding. \citet{seesaghur2023assessment} assessed the use of bone-targeting agents in patients with bone metastases from breast, non-small cell lung, and prostate cancers using EHRs. \citet{morin2021artificial} developed MEDomics, an artificial intelligence framework integrating longitudinal EHRs with real-world data for continuous pan-cancer prognostication, analyzing breast and lung cancer patients. In addition, symptom analysis and patient outcomes were the focus of several studies. \citet{luo2021computational} developed a computational framework to analyze symptom patterns in breast and colorectal cancer patients after chemotherapy. \citet{kehl2021clinical} investigated the use of EHRs to identify cancer patients experiencing clinical inflection points across solid tumors and lymphoma. \citet{luo2021analyzing} analyzed symptoms in colorectal and breast cancer patients, examining the impact of type 2 diabetes comorbidity. \citet{bhatt2023use} investigated the relationship between social support and outcomes in hospitalized patients with advanced cancer.

\citet{zhao2021generating} proposed a framework for generating real-world evidence from unstructured clinical notes to examine the clinical utility of genetic tests in women's cancers. \citet{do2021patterns} investigated patterns of metastatic disease in cancer patients using NLP of structured radiology reports. \citet{shi2022identifying} developed an NLP pipeline to identify patients eligible for genetic testing for hereditary cancers based on family health history data in EHRs. \citet{karimi2021development} created models to identify distant cancer recurrence and recurrence sites for breast cancer and hepatocellular carcinoma patients. \citet{kehl2021artificial} developed deep NLP models to extract clinical outcomes from imaging reports and oncologist notes across various solid tumor types. \citet{chen2023natural} focused on extracting information about esophagitis presence and severity from clinical notes of patients undergoing radiotherapy for lung and esophageal cancers. \citet{huang2023natural} developed an NLP algorithm to extract clinical information from uro-oncological histopathology reports, covering prostate, bladder, and renal cancers. Additionally, few studies explored the extraction of temporal information and social determinants of health. \citet{wang2019achievability} investigated the extraction of specific date information for lung cancer and multiple myeloma cases. \citet{yu2024identifying} developed the SODA package to extract social determinants of health from clinical narratives, examining performance across different race and gender groups and testing generalizability using cancer and opioid use cohorts. \citet{araki2023real} investigated the feasibility of evaluating treatment response in Japanese cancer patients using unstructured EHR data, focusing on lung and breast cancer patients. Furthermore, \citet{huang2024critical} evaluated ChatGPT's ability to extract structured data from clinical notes, focusing on lung cancer and pediatric osteosarcoma pathology reports.

\subsection{Unspecified Cancer Types}
Several studies have explored methods for extracting and classifying cancer-related information, where the specific cancer type was not the primary focus or was unspecified. A rule-based NLP algorithm was developed to automatically extract and categorize tumor response from radiology reports. The proposed algorithm achieved an overall accuracy of 88\% in classifying tumor response as progression or no progression \cite{laurent2023automatic}. Researchers have also created NLP tools to extract specific cancer-related information, such as PD-L1 expression levels from clinical notes \cite{lin2023machine} and physician-reported pain in radiation oncology consultation notes \cite{naseri2021development}. Advanced machine learning techniques, including transformers, have been employed to predict cancer treatment from EHRs \cite{ahmad2023bir}. Additionally, efforts have been made to improve the robustness of classification models for rare cancer types, with a study developing a novel class-specialized ensemble technique that outperformed other methods for rare cancer classification in pathology reports \cite{de2022class}. A study developed an algorithm to extract eastern cooperative oncology group performance status scores across 21 cancer types, improving score completeness from 60.4\% to 73.2\% with 93\% accuracy when applied to a database of over 480,000 patient records \cite{cohen2023natural}. A study has also explored the use of NLP and recurrent neural networks~(RNN) to identify genomic mutation-associated treatment changes in cancer patient progress notes, with bidirectional long short-term memory~(BiLSTM) performing best at 88.6\% accuracy \cite{guan2019natural}.

\subsection{Adverse Event}
Existing studies have also examined various kinds of adverse events associated with cancer and its treatment. A study on Japanese physicians' progress notes achieved high concordance rates for detecting gastrointestinal symptoms, with 83.5\% accuracy for nausea/vomiting and 97.7\% for diarrhea \cite{mashima2022using}. However, the study noted challenges in distinguishing between current symptoms and past medical history mentions. Venous thromboembolism~(VTE), a significant concern in cancer patients, has been addressed in multiple studies. A study developed a computable phenotype for incident VTE using a combination of ICD/medication-based and NLP-based algorithms \cite{li2022developing}. This approach demonstrated high performance with a weighted sensitivity of 96\% and a positive predictive value of 98\%. Another study focused on predicting VTE recurrence in anticoagulated cancer patients, using various machine-learning algorithms on a large dataset from multiple Spanish hospitals \cite{munoz2023development}. The models achieved moderate predictive performance, with AUROC scores ranging from 0.66 to 0.69.

The extraction of common terminology criteria for adverse events~(CTCAE) symptoms from clinical notes has also been investigated. A study evaluated an NLP pipeline based on Apache clinical text analysis knowledge extraction system~(cTAKES) and reported high accuracy for common symptoms in radiation therapy patients, such as radiation dermatitis, fatigue, and nausea \cite{hong2020natural}. However, the system faced challenges in detecting negated symptoms, highlighting an area for future improvement. Researchers have also explored the generalizability of machine learning methods in detecting adverse drug events~(ADEs) from clinical narratives \cite{zitu2023generalizability}. By comparing various machine learning and deep learning methods across different datasets, the study found that ClinicalBERT performed best, achieving F1-scores of 0.78 and 0.74 when tested on distinct datasets.

\section{Discussion}

\begin{figure}[!t] 
    \centering
    \includegraphics[width=.80\textwidth]{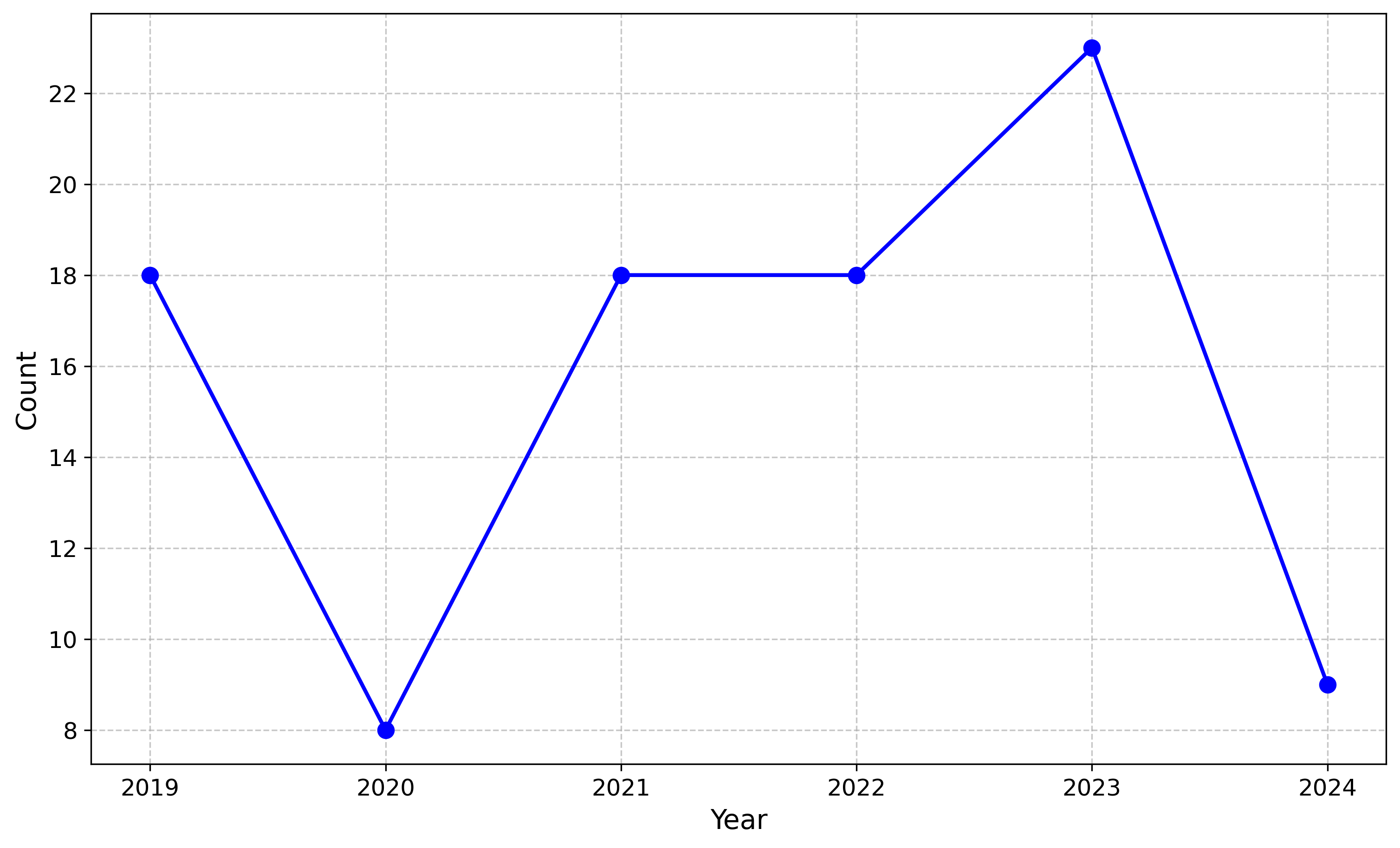}
    \caption{Annual Distribution of Studies (2019-2024).}
    \label{fig:yearly_plot}
\end{figure}

The distribution of studies based on year demonstrates a notable trend over the past six years. As illustrated in Figure~\ref{fig:yearly_plot}, there is a generally upward trajectory in the number of publications from 2019 to 2024. The year 2019 established a solid baseline for subsequent years. 2020 experienced a significant decrease in publications, which can be attributed to the global disruption caused by the COVID-19 pandemic. However, the number of studies increased significantly in 2021, returning to pre-pandemic levels. A marked increase is observed in 2023, indicating growing interest and investment in this research area. Even though the publication data only covers the starting months of 2024, it already indicates encouraging progress and is expected to surpass previous years.

\begin{figure}[!t]
    \centering
    \includegraphics[width=.82\textwidth]{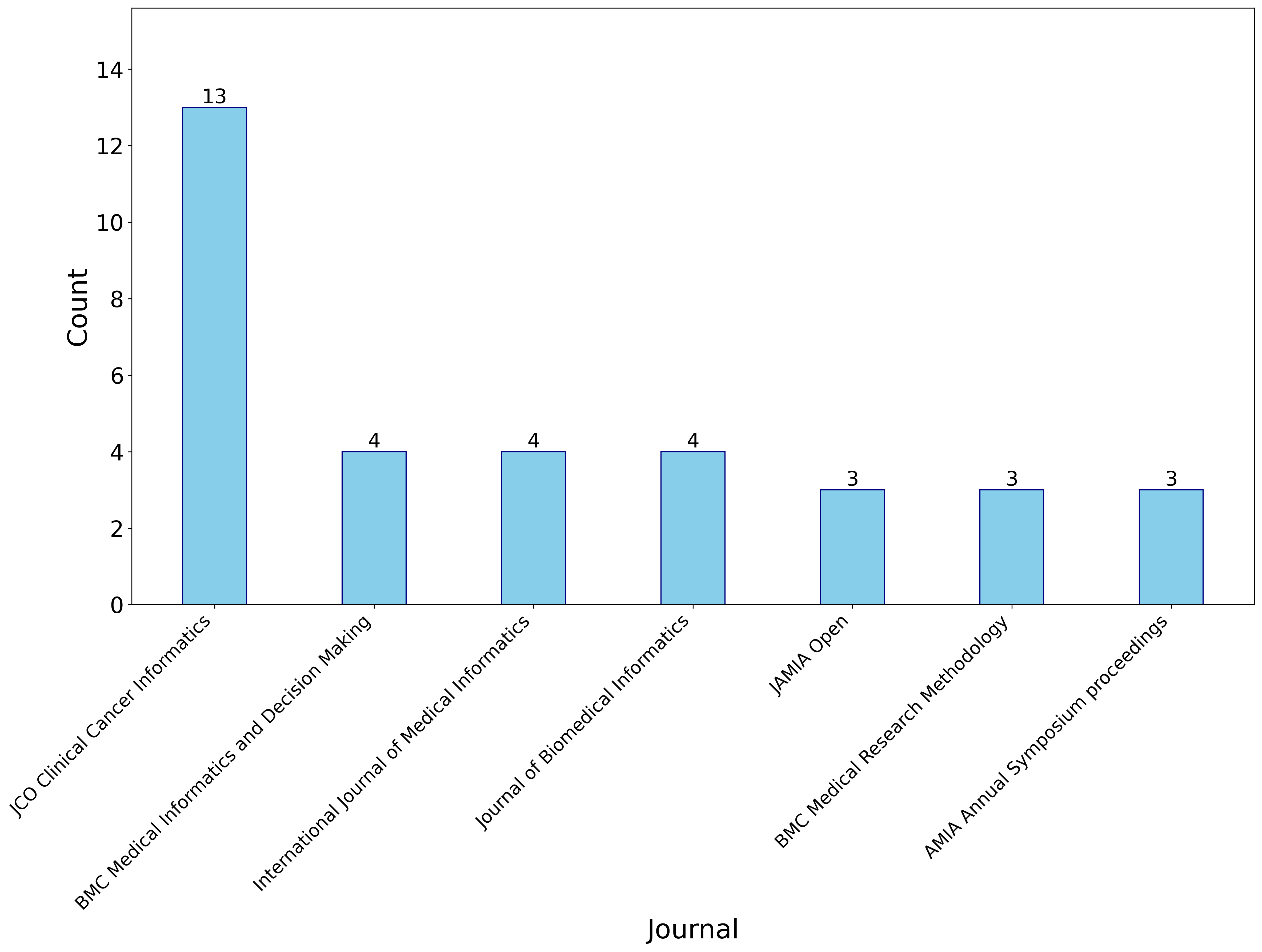}
    \caption{Top journals based on publication count.}
    \label{fig:journal_publication_count}
\end{figure}

\begin{figure}[!t]
    \centering
    \includegraphics[width=.82\textwidth]{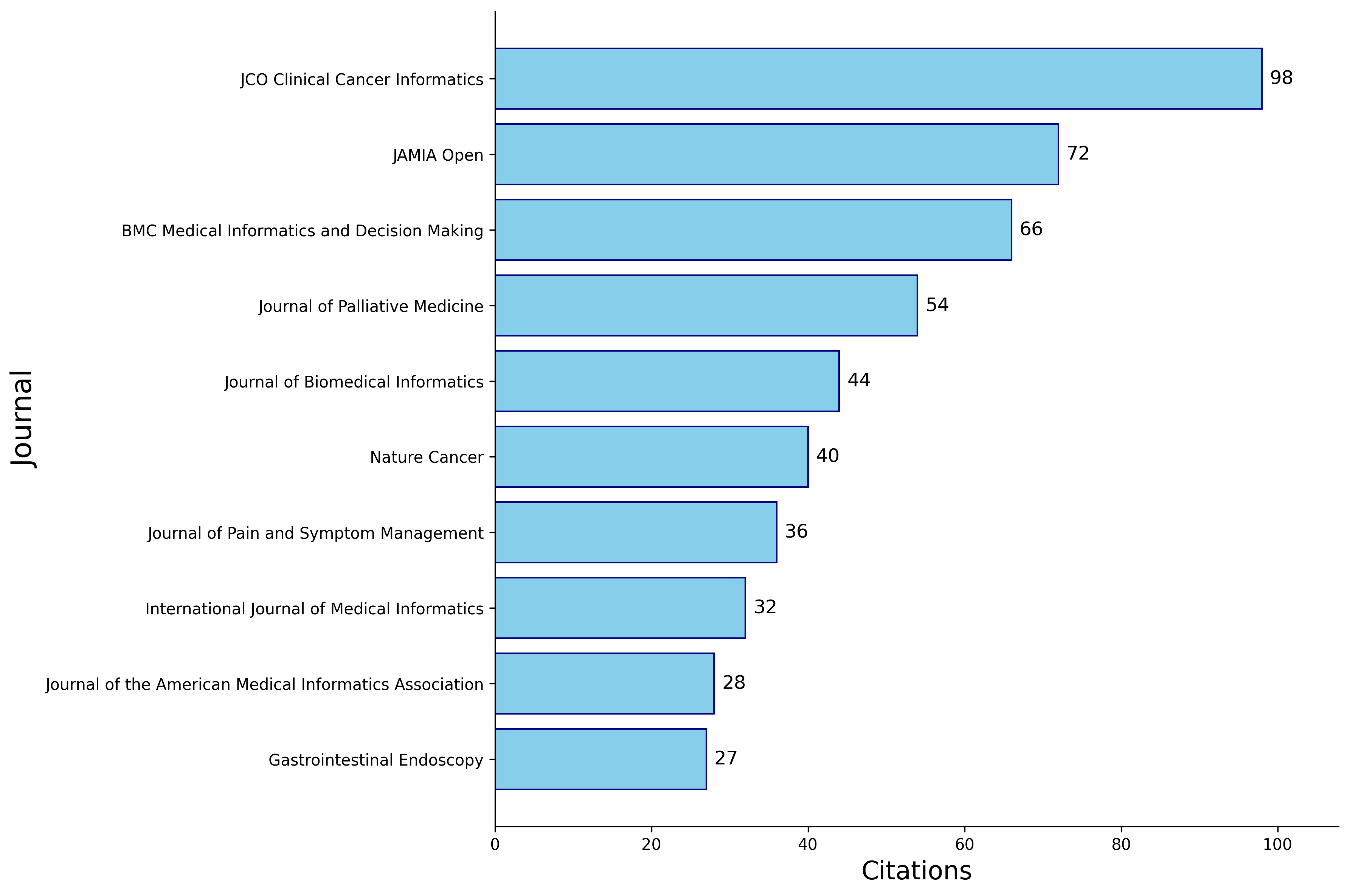}
    \caption{Top 10 journals by citation count.}
    \label{fig:journal_citation_count}
\end{figure}

Figure~\ref{fig:journal_publication_count} illustrates the distribution of publications on NLP for cancer research across top journals from 2019 to 2024. JCO Clinical Cancer Informatics leads with 13 publications, demonstrating its significant role in this field. Three journals follow with 4 publications each: BMC Medical Informatics and Decision Making, International Journal of Medical Informatics, and Journal of Biomedical Informatics. JAMIA Open, BMC Medical Research Methodology, and AMIA Annual Symposium proceedings each contribute 3 publications. Figure~\ref{fig:journal_citation_count} illustrates the Top 10 journals by citation count based on studies included in this review.

\begin{table}[!h]
\centering
\caption{Top 10 most cited papers.}
\label{tab:top10}
\resizebox{\columnwidth}{!}{%
\begin{tabular}{|l|l|l|l|}
\hline
\textbf{Ref.} &
  \textbf{Title} &
  \textbf{Year} &
  \textbf{Citations} \\ \hline
\cite{lindvall2019natural} &
  \begin{tabular}[c]{@{}l@{}}``Natural Language Processing to Assess End-of-Life Quality \\ Indicators in Cancer Patients Receiving Palliative Surgery''\end{tabular} &
  2019 &
  54 \\ \hline
\cite{morin2021artificial} &
  \begin{tabular}[c]{@{}l@{}}``An artificial intelligence framework integrating longitudinal \\ electronic health records with real-world data enables continuous\\ pan-cancer prognostication''\end{tabular} &
  2021 &
  40 \\ \hline
\cite{zhu2019automatically} &
  \begin{tabular}[c]{@{}l@{}}``Automatically identifying social isolation from clinical \\ narratives for patients with prostate Cancer''\end{tabular} &
  2019 &
  32 \\ \hline
\cite{meystre2019automatic} &
  \begin{tabular}[c]{@{}l@{}}``Automatic trial eligibility surveillance based on unstructured\\ clinical data''\end{tabular} &
  2019 &
  31 \\ \hline
\cite{zhou2022cancerbert} &
  \begin{tabular}[c]{@{}l@{}}``CancerBERT: A cancer domain-specific language model for\\ extracting breast cancer phenotypes from electronic health records''\end{tabular} &
  2022 &
  28 \\ \hline
\cite{lindvall2022natural} &
  \begin{tabular}[c]{@{}l@{}}``Natural Language Processing to Identify Advance Care Planning\\ Documentation in a Multisite Pragmatic Clinical Trial''\end{tabular} &
  2022 &
  28 \\ \hline
\cite{laique2021application} &
  \begin{tabular}[c]{@{}l@{}}``Application of optical character recognition with natural language\\ processing for large-scale quality metric data extraction in \\ colonoscopy reports''\end{tabular} &
  2021 &
  27 \\ \hline
\cite{banerjee2019weakly} &
  \begin{tabular}[c]{@{}l@{}}``Weakly supervised natural language processing for assessing\\ patient-centered outcome following   prostate cancer treatment''\end{tabular} &
  2019 &
  26 \\ \hline
\cite{agaronnik2020challenges} &
  \begin{tabular}[c]{@{}l@{}}``Challenges of Developing a Natural Language Processing Method\\ With Electronic Health Records to Identify Persons With Chronic \\ Mobility Disability''\end{tabular} &
  2020 &
  25 \\ \hline
\cite{de2022class} &
  \begin{tabular}[c]{@{}l@{}}``Class imbalance in out-of-distribution datasets: Improving the\\ robustness of the TextCNN for the classification of rare cancer types''\end{tabular} &
  2022 &
  25 \\ \hline
\end{tabular}%
}
\end{table}

Table~\ref{tab:top10} presents the top 10 most cited papers in the field of NLP for cancer research. These citation data reflects the number of citations up to the first quarter of 2024. The most cited paper \cite{lindvall2019natural}, published in 2019, has accumulated 54 citations and explores the use of NLP to assess end-of-life quality indicators in cancer patients undergoing palliative surgery. Other highly cited works include studies on pan-cancer prognostication \cite{morin2021artificial}, social isolation identification \cite{zhu2019automatically}, and automatic trial eligibility surveillance \cite{meystre2019automatic}. This also highlights emerging research areas such as cancer-specific language models \cite{zhou2022cancerbert} and the application of NLP in identifying advance care planning documentation \cite{lindvall2022natural}.

\begin{figure}[!t]
    \centering
    \includegraphics[width=.99\textwidth]{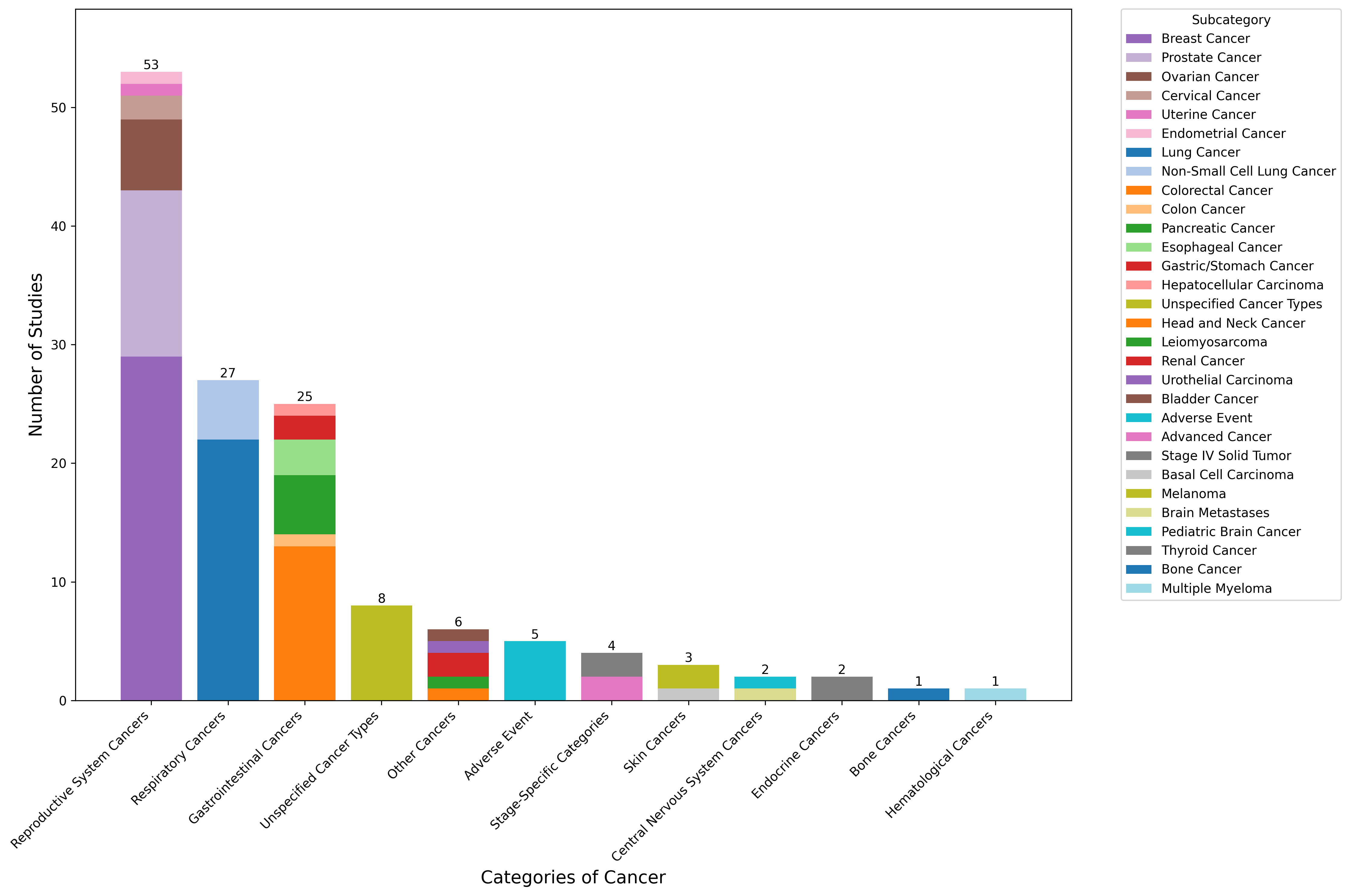}
    \caption{Distribution of Cancer Types in NLP-based Cancer Studies.}
    \label{fig:cancer_studies_chart}
\end{figure}

The analysis of NLP-based studies utilizing EHRs and clinical notes reveals a diverse distribution of cancer types, with some categories receiving more attention than others. Respiratory cancers, particularly lung cancer, were prominently featured in 22 studies, with an additional 5 studies specifically focusing on NSCLC. Gastrointestinal cancers were also well-represented, with colorectal cancer being the subject of 13 studies, followed by pancreatic cancer (5 studies), and esophageal cancer (3 studies). Among reproductive system cancers, breast cancer emerged as the most frequently studied type, with 29 studies dedicated to this area. Prostate cancer and ovarian cancer followed with 14 and 6 studies, respectively. Other cancer types, such as endocrine, skin, CNS, hematological, and bone cancers, were the focus of fewer studies, typically ranging from 1~to~2 per specific subtype. The analysis also included studies on stage-specific categories, unspecified cancer types, and adverse events related to cancer treatment. This distribution presented in Figure~\ref{fig:cancer_studies_chart} highlights the current research priorities in applying NLP techniques to cancer-related EHRs and clinical notes, while also indicating potential areas for future exploration in less-studied cancer types.

\section{NLP Applications and Methodologies in Cancer Research}
The applications of NLP in cancer research are diverse and wide-ranging. Common tasks include information extraction, where specific entities and attributes related to cancer diagnosis, staging, treatment, and outcomes are retrieved from clinical notes and pathology reports \cite{hong2020annotation,hochheiser2023deepphe,zhou2022cancerbert,ali2022development,yoo2022transforming}. For example, existing studies have used NLP to extract tumor characteristics from radiology reports \cite{pathak2023extracting} and PD-L1 expression levels from clinical notes \cite{lin2023machine}. Text classification is another key application, involving the categorization of clinical documents into predefined groups such as cancer stage \cite{macchia2022multidisciplinary,bozkurt2022expanding}, smoking status \cite{ebrahimi2024identification}, or presence of metastasis \cite{li2023interpretable,yang2022identification}.

Named Entity Recognition (NER) is frequently employed to identify and classify entities like cancer types, medications, and symptoms in clinical texts \cite{garcia2024gpt,paul2022investigation}. Relation extraction techniques are used to detect relationships between entities, such as associations between symptoms and patient attributes \cite{luo2021computational}. NLP has also been crucial in phenotyping, which involves automatically identifying patient cohorts with specific clinical characteristics from EHR data \cite{li2022developing,ernecoff2019electronic,liu2023leveraging}. Additionally, predictive modeling leveraging NLP-derived features has been used to forecast cancer outcomes, recurrence risk, and treatment response \cite{sanyal2021weakly,kehl2021clinical,laios2023roberta}.

The methodologies used in these NLP applications span a wide spectrum, from traditional rule-based approaches to cutting-edge machine learning techniques. Many studies, especially those focused on well-defined extraction tasks, have utilized rule-based NLP approaches \cite{hong2020annotation,alba2021ascertainment,lindvall2022natural,ali2022development}. These methods rely on predefined rules, regular expressions, and dictionaries to identify relevant information. Traditional machine learning algorithms such as support vector machine~(SVM), Random Forest, and Logistic Regression have been applied to various classification and extraction tasks \cite{yang2022identification,ribelles2021machine,zeng2019identifying}.

In recent years, there has been a shift towards deep learning architectures. CNN \cite{harrison2023successful,wang2019natural}, RNN \cite{sanyal2021weakly}, and long short-term memory~(LSTM) \cite{bozkurt2020phenotyping,guan2019natural} have shown improved performance on complex NLP tasks. State-of-the-art transformer-based models like BERT, RoBERTa, and their domain-specific variants (e.g., ClinicalBERT, CancerBERT) have demonstrated particularly promising results across various cancer-related NLP tasks \cite{li2023interpretable,zhou2022cancerbert,solarte2023transformers,pathak2023extracting,laios2023roberta}. Some researchers have also explored hybrid approaches, combining rule-based methods with machine learning techniques to leverage the strengths of both \cite{solarte2021integrating,bozkurt2022expanding,bozkurt2020phenotyping}.

Text preprocessing and feature extraction play crucial roles in the NLP pipeline for cancer research. Common preprocessing steps include tokenization \cite{hong2020annotation,ruckdeschel2023unstructured,hochheiser2023deepphe,pathak2023extracting}, sentence boundary detection \cite{pathak2023extracting,ebrahimi2024identification}, text cleaning and normalization \cite{bozkurt2022expanding,chen2022automated,ebrahimi2024identification}, and removal of stop words and punctuation \cite{ebrahimi2024identification,bozkurt2019possible}. Feature extraction techniques range from traditional bag-of-words~(BoW) and term frequency–inverse document frequency~(TF-IDF) representations \cite{yang2022identification,ribelles2021machine,morin2021artificial} to more advanced word embeddings like Word2Vec and GloVe \cite{zhou2022cancerbert,sanyal2021weakly,bozkurt2022expanding}, and contextual embeddings derived from transformer models \cite{li2023interpretable,zhou2022cancerbert}.

The performance of these NLP models is typically assessed using standard metrics such as precision, recall, and F1-score \cite{hong2020annotation,li2023interpretable,hochheiser2023deepphe,zhou2022cancerbert,lindvall2022natural,ali2022development}, accuracy \cite{macchia2022multidisciplinary,laique2021application,laurent2023automatic}, AUROC \cite{ribelles2021machine,sanyal2021weakly,karimi2021development}, and sensitivity and specificity \cite{yang2022identification,ernecoff2019electronic,liu2023leveraging}. Some studies also report additional metrics like Cohen's Kappa \cite{huang2024critical} for inter-rater agreement or mean absolute error~(MAE) \cite{lin2023machine} for regression tasks.

\section{NLP-based Datasets in Cancer Research}
NLP has become an essential tool in cancer research, leveraging vast amounts of unstructured clinical data to extract valuable insights. The primary source of data for most NLP studies in oncology is EHRs. These datasets vary widely in size and scope, ranging from institutional databases to multi-center collaborations and national registries. Many studies utilize EHRs from specific hospitals or healthcare systems. For example, the University of Florida Health Integrated Data Repository provided over 1.5 million clinical notes from 20,000 cancer patients for several studies \cite{yu2021study, yu2024identifying}. Similarly, Mayo Clinic EHRs containing records of 4,110 lung cancer patients \cite{wang2019achievability} and Stanford's prostate cancer research database with 528,362 notes from 6,595 patients \cite{banerjee2019weakly} have been instrumental in advancing NLP research in oncology. Larger, multi-institutional databases have also played a crucial role. The Flatiron Health database, which aggregates EHRs from approximately 280 US cancer clinics and contains data on over 300,000 cancer patients, has been used in multiple studies \cite{cohen2023natural, adamson2023approach, benedum2023replication}. Another notable resource is the MIMIC-III dataset, which includes records of 46,146 patients \cite{ruckdeschel2023unstructured}. Cancer registry data, such as those from the Kentucky Cancer Registry and Louisiana Tumor Registry, have been particularly useful for extracting information from pathology reports \cite{hong2020annotation, de2022class}.

The size of datasets used in these studies varies considerably. Large-scale studies have employed datasets with hundreds of thousands of patient records, such as one involving 102,475 patients \cite{liu2023leveraging} and another with 186,313 lung cancer patients \cite{benedum2023replication}. Medium-scale studies typically involved thousands to tens of thousands of patients, exemplified by a study using data from 5,152 patients with 366,398 clinical notes \cite{luo2021computational}. Smaller, more focused studies have worked with hundreds of patients such as one that developed NLP systems using data from 200 patients \cite{mashima2022using}. These datasets contains various types of clinical documents, including clinical notes, progress reports \cite{guan2019natural, yu2021study}, radiology reports \cite{yang2022identification, huang2024critical}, pathology reports \cite{harrison2023successful, de2022class}, and discharge summaries \cite{hong2020annotation, zitu2023generalizability}. This diversity allows researchers to extract a wide range of information relevant to cancer diagnosis, treatment, and outcomes. While many of these datasets are not publicly available due to privacy concerns, there is a growing trend towards making more resources accessible to researchers. The MIMIC-III dataset, for example, is publicly available \cite{ruckdeschel2023unstructured}, and I2B2 datasets can be accessed upon request \cite{ahmad2023bir, zitu2023generalizability}. Some institutional datasets may also be available upon request to the authors \cite{jin2021research, chen2022automated}.

\section{Open Research Challenges and Future Directions}
NLP has shown significant potential in advancing cancer research and clinical practice. However, several key challenges and opportunities for future work have emerged. This section provides an overview of key challenges and future directions for the application of NLP techniques in cancer research.

\subsection{Data Limitations and Generalizability}
Many studies are limited by small sample sizes, single-institution data, or focus on specific cancer types, limiting the generalizability of NLP models \cite{zitu2023generalizability, li2023interpretable, mashima2022using, hochheiser2023deepphe, zhou2023cross, zhou2022cancerbert, alba2021ascertainment, lindvall2022natural, bozkurt2022expanding, ali2022development, hong2020natural}. This restriction in data scope presents a significant challenge to the broad application of NLP in oncology. To address this, future work should focus on validating NLP algorithms across multiple institutions and diverse patient populations \cite{zhou2023cross, zhou2022cancerbert, hernandez2020leveraging, bozkurt2022expanding, morin2021artificial, do2021patterns}. This multi-institutional approach will help ensure that the developed models are robust and applicable in various clinical settings. Additionally, expanding datasets to include more cancer types and clinical scenarios \cite{garcia2024gpt, liu2023leveraging, pathak2023extracting, huang2023natural} will enhance the versatility of NLP systems in oncology. Researchers should also develop methods to handle variations in clinical documentation practices across institutions \cite{gauthier2022automating, ernecoff2019electronic}, as these differences can significantly impact the performance of NLP models.

\subsection{Model Performance and Robustness}
Current NLP models face challenges in handling complex clinical language, negations, and context. These limitations can lead to misinterpretation of clinical notes and potentially impact patient care. To improve model performance and robustness, future research should focus on enhancing algorithms to better handle variability in report formats and vocabularies \cite{laurent2023automatic, dimartino2022identification, chen2022automated}. This includes improving negation and speculation detection \cite{solarte2021integrating, hong2020natural}, which is crucial for accurately interpreting clinical narratives. Furthermore, developing more advanced NLP techniques for automated symptom extraction \cite{luo2021computational, chen2023natural} will enable a more detailed analysis of patient experiences and outcomes.

\subsection{Integration with Clinical Workflows}
To maximize impact, NLP systems need seamless integration into clinical practice. This integration is essential for the practical application of NLP in real-world healthcare settings. Future efforts should focus on developing workflow-embedded clinical decision support tools \cite{liu2023leveraging, karwa2020development} that can provide real-time insights to clinicians. Creating user-friendly interfaces for database development and querying \cite{harrison2023successful} will make NLP tools more accessible to healthcare professionals who may not have extensive technical expertise. Additionally, implementing real-time NLP systems for continuous patient monitoring \cite{morin2021artificial} can help in the early detection of adverse events and improve patient care.

\subsection{Expanding NLP Applications}
There are numerous opportunities to apply NLP to new areas of cancer research. Extracting information from multimodal data sources (e.g., imaging reports, genomic data) \cite{alba2021ascertainment, adamson2023approach, guan2019natural} can provide a more comprehensive view of patient health. Analyzing social determinants of health in cancer patients \cite{yu2021study, yu2024identifying, yu2022assessing} through NLP can help identify factors influencing treatment outcomes and patient experiences. Automating clinical trial eligibility screening \cite{meystre2019automatic} using NLP can accelerate patient recruitment and improve the efficiency of clinical research.

\subsection{Ethical and Privacy Considerations}
As NLP systems handle sensitive patient data, addressing privacy and ethical concerns is crucial. Future research should focus on developing privacy-preserving NLP techniques \cite{hu2024zero, hernandez2020leveraging} to ensure patient confidentiality while leveraging the power of large datasets. Ensuring fairness and reducing bias in NLP models \cite{hernandez2020leveraging, yu2024identifying} is also essential to prevent disparities in healthcare delivery and research outcomes.

\subsection{Technical Advancements}
Several technical areas require further research to enhance the capabilities of NLP in cancer research. This includes exploring advanced deep learning architectures like transformers and large language models~(LLMs) \cite{huang2024critical, adamson2023approach, senders2019natural}, which have shown promise in handling complex language tasks. Improving the interpretability of NLP models \cite{li2023interpretable, huang2024critical} is crucial for gaining clinician trust and understanding model decision-making processes. Developing hybrid rule-based and machine learning approaches \cite{hong2020annotation, solarte2021integrating} can combine the strengths of both methodologies to create more robust and flexible NLP systems. Additionally, the potential of LLMs in advancing cancer research through improved text analysis and knowledge extraction is an emerging area for future research \cite{iannantuono2023applications}.

\section{Conclusion}

This review examines the current state-of-the-art literature on NLP applications in analyzing EHRs and clinical notes for cancer research. The analysis of 94 studies published since 2019 shows a clear trend in the growth of this field, with a notable increase in publications in 2023. The distribution of studies across cancer types revealed a concentration on specific areas. Breast cancer was the most studied by existing studies, followed by lung cancer, and colorectal cancer. This focus likely reflects both the prevalence of these cancers and the availability of data. However, it also highlights potential gaps in research for less common cancer types, which could benefit from increased attention in future studies. In terms of NLP applications, information extraction and text classification emerged as the most common tasks. Specifically, studies focused on extracting tumor characteristics from radiology reports, PD-L1 expression levels from clinical notes, and classifying cancer stages and metastasis presence. The NLP techniques used have evolved from rule-based approaches to more sophisticated machine learning techniques, with a notable trend towards the use of transformer-based models like BERT and its variants (e.g., CancerBERT).

While NLP has shown significant promise in cancer research using EHRs and clinical notes, there remains substantial room for improvement in model generalizability, performance, and clinical integration. Additionally, expanding NLP applications to analyze social determinants of health in cancer patients and automating clinical trial eligibility screening represent promising avenues for enhancing patient care and accelerating research. The potential of large language models in advancing cancer research, as suggested by recent studies, also presents an exciting frontier for future exploration. Addressing these challenges will be crucial to revolutionizing cancer research and patient care, leading to improved diagnosis, treatment, outcomes, and more personalized treatment approaches.

\section*{Declaration of Interests}
\noindent The authors have declared no conflict of interest. 

\newpage
\bibliographystyle{elsarticle-num-names} 
\bibliography{ref}

\begin{thebibliography}{112}
\expandafter\ifx\csname natexlab\endcsname\relax\def\natexlab#1{#1}\fi
\providecommand{\url}[1]{\texttt{#1}}
\providecommand{\href}[2]{#2}
\providecommand{\path}[1]{#1}
\providecommand{\DOIprefix}{doi:}
\providecommand{\ArXivprefix}{arXiv:}
\providecommand{\URLprefix}{URL: }
\providecommand{\Pubmedprefix}{pmid:}
\providecommand{\doi}[1]{\href{http://dx.doi.org/#1}{\path{#1}}}
\providecommand{\Pubmed}[1]{\href{pmid:#1}{\path{#1}}}
\providecommand{\bibinfo}[2]{#2}
\ifx\xfnm\relax \def\xfnm[#1]{\unskip,\space#1}\fi
\bibitem[{Siegel et~al.(2024)Siegel, Giaquinto, and Jemal}]{siegel2024cancer}
\bibinfo{author}{R.~L. Siegel}, \bibinfo{author}{A.~N. Giaquinto}, \bibinfo{author}{A.~Jemal},
\newblock \bibinfo{title}{Cancer statistics, 2024.},
\newblock \bibinfo{journal}{CA: a cancer journal for clinicians} \bibinfo{volume}{74} (\bibinfo{year}{2024}).
\bibitem[{Schepis et~al.(2023)Schepis, De~Lucia, Pellegrino, Del~Gaudio, Maresca, Coppola, Chiappetta, Gasbarrini, Franceschi, Candelli et~al.}]{schepis2023state}
\bibinfo{author}{T.~Schepis}, \bibinfo{author}{S.~S. De~Lucia}, \bibinfo{author}{A.~Pellegrino}, \bibinfo{author}{A.~Del~Gaudio}, \bibinfo{author}{R.~Maresca}, \bibinfo{author}{G.~Coppola}, \bibinfo{author}{M.~F. Chiappetta}, \bibinfo{author}{A.~Gasbarrini}, \bibinfo{author}{F.~Franceschi}, \bibinfo{author}{M.~Candelli}, et~al.,
\newblock \bibinfo{title}{State-of-the-art and upcoming innovations in pancreatic cancer care: a step forward to precision medicine},
\newblock \bibinfo{journal}{Cancers} \bibinfo{volume}{15} (\bibinfo{year}{2023}) \bibinfo{pages}{3423}.
\bibitem[{Lu et~al.(2023)Lu, Chen, Liu, Jiao, Liu, Wang, Zhang, Jia, Gong, Yang et~al.}]{lu2023landscape}
\bibinfo{author}{Z.~Lu}, \bibinfo{author}{Y.~Chen}, \bibinfo{author}{D.~Liu}, \bibinfo{author}{X.~Jiao}, \bibinfo{author}{C.~Liu}, \bibinfo{author}{Y.~Wang}, \bibinfo{author}{Z.~Zhang}, \bibinfo{author}{K.~Jia}, \bibinfo{author}{J.~Gong}, \bibinfo{author}{Z.~Yang}, et~al.,
\newblock \bibinfo{title}{The landscape of cancer research and cancer care in china},
\newblock \bibinfo{journal}{Nature Medicine} \bibinfo{volume}{29} (\bibinfo{year}{2023}) \bibinfo{pages}{3022--3032}.
\bibitem[{Shmatko et~al.(2022)Shmatko, Ghaffari~Laleh, Gerstung, and Kather}]{shmatko2022artificial}
\bibinfo{author}{A.~Shmatko}, \bibinfo{author}{N.~Ghaffari~Laleh}, \bibinfo{author}{M.~Gerstung}, \bibinfo{author}{J.~N. Kather},
\newblock \bibinfo{title}{Artificial intelligence in histopathology: enhancing cancer research and clinical oncology},
\newblock \bibinfo{journal}{Nature cancer} \bibinfo{volume}{3} (\bibinfo{year}{2022}) \bibinfo{pages}{1026--1038}.
\bibitem[{Veta et~al.(2014)Veta, Pluim, Van~Diest, and Viergever}]{veta2014breast}
\bibinfo{author}{M.~Veta}, \bibinfo{author}{J.~P. Pluim}, \bibinfo{author}{P.~J. Van~Diest}, \bibinfo{author}{M.~A. Viergever},
\newblock \bibinfo{title}{Breast cancer histopathology image analysis: A review},
\newblock \bibinfo{journal}{IEEE transactions on biomedical engineering} \bibinfo{volume}{61} (\bibinfo{year}{2014}) \bibinfo{pages}{1400--1411}.
\bibitem[{Bi et~al.(2019)Bi, Hosny, Schabath, Giger, Birkbak, Mehrtash, Allison, Arnaout, Abbosh, Dunn et~al.}]{bi2019artificial}
\bibinfo{author}{W.~L. Bi}, \bibinfo{author}{A.~Hosny}, \bibinfo{author}{M.~B. Schabath}, \bibinfo{author}{M.~L. Giger}, \bibinfo{author}{N.~J. Birkbak}, \bibinfo{author}{A.~Mehrtash}, \bibinfo{author}{T.~Allison}, \bibinfo{author}{O.~Arnaout}, \bibinfo{author}{C.~Abbosh}, \bibinfo{author}{I.~F. Dunn}, et~al.,
\newblock \bibinfo{title}{Artificial intelligence in cancer imaging: clinical challenges and applications},
\newblock \bibinfo{journal}{CA: a cancer journal for clinicians} \bibinfo{volume}{69} (\bibinfo{year}{2019}) \bibinfo{pages}{127--157}.
\bibitem[{Elemento et~al.(2021)Elemento, Leslie, Lundin, and Tourassi}]{elemento2021artificial}
\bibinfo{author}{O.~Elemento}, \bibinfo{author}{C.~Leslie}, \bibinfo{author}{J.~Lundin}, \bibinfo{author}{G.~Tourassi},
\newblock \bibinfo{title}{Artificial intelligence in cancer research, diagnosis and therapy},
\newblock \bibinfo{journal}{Nature Reviews Cancer} \bibinfo{volume}{21} (\bibinfo{year}{2021}) \bibinfo{pages}{747--752}.
\bibitem[{Sitapati et~al.(2017)Sitapati, Kim, Berkovich, Marmor, Singh, El-Kareh, Clay, and Ohno-Machado}]{sitapati2017integrated}
\bibinfo{author}{A.~Sitapati}, \bibinfo{author}{H.~Kim}, \bibinfo{author}{B.~Berkovich}, \bibinfo{author}{R.~Marmor}, \bibinfo{author}{S.~Singh}, \bibinfo{author}{R.~El-Kareh}, \bibinfo{author}{B.~Clay}, \bibinfo{author}{L.~Ohno-Machado},
\newblock \bibinfo{title}{Integrated precision medicine: the role of electronic health records in delivering personalized treatment},
\newblock \bibinfo{journal}{Wiley Interdisciplinary Reviews: Systems Biology and Medicine} \bibinfo{volume}{9} (\bibinfo{year}{2017}) \bibinfo{pages}{e1378}.
\bibitem[{Jensen et~al.(2012)Jensen, Jensen, and Brunak}]{jensen2012mining}
\bibinfo{author}{P.~B. Jensen}, \bibinfo{author}{L.~J. Jensen}, \bibinfo{author}{S.~Brunak},
\newblock \bibinfo{title}{Mining electronic health records: towards better research applications and clinical care},
\newblock \bibinfo{journal}{Nature Reviews Genetics} \bibinfo{volume}{13} (\bibinfo{year}{2012}) \bibinfo{pages}{395--405}.
\bibitem[{Kreimeyer et~al.(2017)Kreimeyer, Foster, Pandey, Arya, Halford, Jones, Forshee, Walderhaug, and Botsis}]{kreimeyer2017natural}
\bibinfo{author}{K.~Kreimeyer}, \bibinfo{author}{M.~Foster}, \bibinfo{author}{A.~Pandey}, \bibinfo{author}{N.~Arya}, \bibinfo{author}{G.~Halford}, \bibinfo{author}{S.~F. Jones}, \bibinfo{author}{R.~Forshee}, \bibinfo{author}{M.~Walderhaug}, \bibinfo{author}{T.~Botsis},
\newblock \bibinfo{title}{Natural language processing systems for capturing and standardizing unstructured clinical information: a systematic review},
\newblock \bibinfo{journal}{Journal of biomedical informatics} \bibinfo{volume}{73} (\bibinfo{year}{2017}) \bibinfo{pages}{14--29}.
\bibitem[{Tayefi et~al.(2021)Tayefi, Ngo, Chomutare, Dalianis, Salvi, Budrionis, and Godtliebsen}]{tayefi2021challenges}
\bibinfo{author}{M.~Tayefi}, \bibinfo{author}{P.~Ngo}, \bibinfo{author}{T.~Chomutare}, \bibinfo{author}{H.~Dalianis}, \bibinfo{author}{E.~Salvi}, \bibinfo{author}{A.~Budrionis}, \bibinfo{author}{F.~Godtliebsen},
\newblock \bibinfo{title}{Challenges and opportunities beyond structured data in analysis of electronic health records},
\newblock \bibinfo{journal}{Wiley Interdisciplinary Reviews: Computational Statistics} \bibinfo{volume}{13} (\bibinfo{year}{2021}) \bibinfo{pages}{e1549}.
\bibitem[{Datta et~al.(2019)Datta, Bernstam, and Roberts}]{datta2019frame}
\bibinfo{author}{S.~Datta}, \bibinfo{author}{E.~V. Bernstam}, \bibinfo{author}{K.~Roberts},
\newblock \bibinfo{title}{A frame semantic overview of nlp-based information extraction for cancer-related ehr notes},
\newblock \bibinfo{journal}{Journal of biomedical informatics} \bibinfo{volume}{100} (\bibinfo{year}{2019}) \bibinfo{pages}{103301}.
\bibitem[{Aggarwal and Pallavi(2024)}]{aggarwal2024advancements}
\bibinfo{author}{D.~Aggarwal}, \bibinfo{author}{K.~Pallavi},
\newblock \bibinfo{title}{Advancements and challenges in natural language processing in oral cancer research: A narrative review},
\newblock \bibinfo{journal}{Cancer Research, Statistics, and Treatment} \bibinfo{volume}{7} (\bibinfo{year}{2024}) \bibinfo{pages}{228--233}.
\bibitem[{Wang et~al.(2022)Wang, Fu, Wen, Ruan, He, Liu, Moon, Mai, Riaz, Wang et~al.}]{wang2022assessment}
\bibinfo{author}{L.~Wang}, \bibinfo{author}{S.~Fu}, \bibinfo{author}{A.~Wen}, \bibinfo{author}{X.~Ruan}, \bibinfo{author}{H.~He}, \bibinfo{author}{S.~Liu}, \bibinfo{author}{S.~Moon}, \bibinfo{author}{M.~Mai}, \bibinfo{author}{I.~B. Riaz}, \bibinfo{author}{N.~Wang}, et~al.,
\newblock \bibinfo{title}{Assessment of electronic health record for cancer research and patient care through a scoping review of cancer natural language processing},
\newblock \bibinfo{journal}{JCO Clinical Cancer Informatics} \bibinfo{volume}{6} (\bibinfo{year}{2022}) \bibinfo{pages}{e2200006}.
\bibitem[{Gholipour et~al.(2023)Gholipour, Khajouei, Amiri, Hajesmaeel~Gohari, and Ahmadian}]{gholipour2023extracting}
\bibinfo{author}{M.~Gholipour}, \bibinfo{author}{R.~Khajouei}, \bibinfo{author}{P.~Amiri}, \bibinfo{author}{S.~Hajesmaeel~Gohari}, \bibinfo{author}{L.~Ahmadian},
\newblock \bibinfo{title}{Extracting cancer concepts from clinical notes using natural language processing: a systematic review},
\newblock \bibinfo{journal}{BMC bioinformatics} \bibinfo{volume}{24} (\bibinfo{year}{2023}) \bibinfo{pages}{405}.
\bibitem[{Bitterman et~al.(2021)Bitterman, Miller, Mak, and Savova}]{bitterman2021clinical}
\bibinfo{author}{D.~S. Bitterman}, \bibinfo{author}{T.~A. Miller}, \bibinfo{author}{R.~H. Mak}, \bibinfo{author}{G.~K. Savova},
\newblock \bibinfo{title}{Clinical natural language processing for radiation oncology: a review and practical primer},
\newblock \bibinfo{journal}{International Journal of Radiation Oncology* Biology* Physics} \bibinfo{volume}{110} (\bibinfo{year}{2021}) \bibinfo{pages}{641--655}.
\bibitem[{Sangariyavanich et~al.(2023)Sangariyavanich, Ponthongmak, Tansawet, Theera-Ampornpunt, Numthavaj, McKay, Attia, and Thakkinstian}]{sangariyavanich2023systematic}
\bibinfo{author}{E.~Sangariyavanich}, \bibinfo{author}{W.~Ponthongmak}, \bibinfo{author}{A.~Tansawet}, \bibinfo{author}{N.~Theera-Ampornpunt}, \bibinfo{author}{P.~Numthavaj}, \bibinfo{author}{G.~J. McKay}, \bibinfo{author}{J.~Attia}, \bibinfo{author}{A.~Thakkinstian},
\newblock \bibinfo{title}{Systematic review of natural language processing for recurrent cancer detection from electronic medical records},
\newblock \bibinfo{journal}{Informatics in Medicine Unlocked}  (\bibinfo{year}{2023}) \bibinfo{pages}{101326}.
\bibitem[{Hong et~al.(2020)Hong, Davoudi, Yu, and Mowery}]{hong2020annotation}
\bibinfo{author}{J.~Hong}, \bibinfo{author}{A.~Davoudi}, \bibinfo{author}{S.~Yu}, \bibinfo{author}{D.~L. Mowery},
\newblock \bibinfo{title}{Annotation and extraction of age and temporally-related events from clinical histories},
\newblock \bibinfo{journal}{BMC Medical Informatics and Decision Making} \bibinfo{volume}{20} (\bibinfo{year}{2020}) \bibinfo{pages}{1--15}.
\bibitem[{Ruckdeschel et~al.(2023)Ruckdeschel, Riley, Parsatharathy, Chamarthi, Rajagopal, Hsu, Mangold, and Driscoll}]{ruckdeschel2023unstructured}
\bibinfo{author}{J.~C. Ruckdeschel}, \bibinfo{author}{M.~Riley}, \bibinfo{author}{S.~Parsatharathy}, \bibinfo{author}{R.~Chamarthi}, \bibinfo{author}{C.~Rajagopal}, \bibinfo{author}{H.~S. Hsu}, \bibinfo{author}{D.~Mangold}, \bibinfo{author}{C.~Driscoll},
\newblock \bibinfo{title}{Unstructured data are superior to structured data for eliciting quantitative smoking history from the electronic health record},
\newblock \bibinfo{journal}{JCO Clinical Cancer Informatics} \bibinfo{volume}{7} (\bibinfo{year}{2023}) \bibinfo{pages}{e2200155}.
\bibitem[{Hu et~al.(2024)Hu, Liu, Zhu, Lu, and Wu}]{hu2024zero}
\bibinfo{author}{D.~Hu}, \bibinfo{author}{B.~Liu}, \bibinfo{author}{X.~Zhu}, \bibinfo{author}{X.~Lu}, \bibinfo{author}{N.~Wu},
\newblock \bibinfo{title}{Zero-shot information extraction from radiological reports using chatgpt},
\newblock \bibinfo{journal}{International Journal of Medical Informatics} \bibinfo{volume}{183} (\bibinfo{year}{2024}) \bibinfo{pages}{105321}.
\bibitem[{Solarte~Pabon et~al.(2021)Solarte~Pabon, Torrente, Provencio, Rodr{\'\i}guez-Gonzalez, and Menasalvas}]{solarte2021integrating}
\bibinfo{author}{O.~Solarte~Pabon}, \bibinfo{author}{M.~Torrente}, \bibinfo{author}{M.~Provencio}, \bibinfo{author}{A.~Rodr{\'\i}guez-Gonzalez}, \bibinfo{author}{E.~Menasalvas},
\newblock \bibinfo{title}{Integrating speculation detection and deep learning to extract lung cancer diagnosis from clinical notes},
\newblock \bibinfo{journal}{Applied Sciences} \bibinfo{volume}{11} (\bibinfo{year}{2021}) \bibinfo{pages}{865}.
\bibitem[{Gauthier et~al.(2022)Gauthier, Law, Le, Li, Zahir, Nirmalakumar, Sung, Pettengell, Aviv, Chu et~al.}]{gauthier2022automating}
\bibinfo{author}{M.-P. Gauthier}, \bibinfo{author}{J.~H. Law}, \bibinfo{author}{L.~W. Le}, \bibinfo{author}{J.~J. Li}, \bibinfo{author}{S.~Zahir}, \bibinfo{author}{S.~Nirmalakumar}, \bibinfo{author}{M.~Sung}, \bibinfo{author}{C.~Pettengell}, \bibinfo{author}{S.~Aviv}, \bibinfo{author}{R.~Chu}, et~al.,
\newblock \bibinfo{title}{Automating access to real-world evidence},
\newblock \bibinfo{journal}{JTO Clinical and Research Reports} \bibinfo{volume}{3} (\bibinfo{year}{2022}) \bibinfo{pages}{100340}.
\bibitem[{Liu et~al.(2023)Liu, McCoy, Aldrich, Sandler, Reese, Steitz, Bian, Wu, Russo, and Wright}]{liu2023leveraging}
\bibinfo{author}{S.~Liu}, \bibinfo{author}{A.~B. McCoy}, \bibinfo{author}{M.~C. Aldrich}, \bibinfo{author}{K.~L. Sandler}, \bibinfo{author}{T.~J. Reese}, \bibinfo{author}{B.~Steitz}, \bibinfo{author}{J.~Bian}, \bibinfo{author}{Y.~Wu}, \bibinfo{author}{E.~Russo}, \bibinfo{author}{A.~Wright},
\newblock \bibinfo{title}{Leveraging natural language processing to identify eligible lung cancer screening patients with the electronic health record},
\newblock \bibinfo{journal}{International Journal of Medical Informatics} \bibinfo{volume}{177} (\bibinfo{year}{2023}) \bibinfo{pages}{105136}.
\bibitem[{Yang et~al.(2024)Yang, Yang, Lyu, Huang, Chen, He, Braithwaite, Mehta, Wu, Guo et~al.}]{yang2024extracting}
\bibinfo{author}{S.~Yang}, \bibinfo{author}{X.~Yang}, \bibinfo{author}{T.~Lyu}, \bibinfo{author}{J.~L. Huang}, \bibinfo{author}{A.~Chen}, \bibinfo{author}{X.~He}, \bibinfo{author}{D.~Braithwaite}, \bibinfo{author}{H.~J. Mehta}, \bibinfo{author}{Y.~Wu}, \bibinfo{author}{Y.~Guo}, et~al.,
\newblock \bibinfo{title}{Extracting pulmonary nodules and nodule characteristics from radiology reports of lung cancer screening patients using transformer models},
\newblock \bibinfo{journal}{Journal of Healthcare Informatics Research}  (\bibinfo{year}{2024}) \bibinfo{pages}{1--15}.
\bibitem[{Ebrahimi et~al.(2024)Ebrahimi, Henriksen, Brasen, Hilberg, Hansen, Jensen, Peimankar, and Wiil}]{ebrahimi2024identification}
\bibinfo{author}{A.~Ebrahimi}, \bibinfo{author}{M.~B.~H. Henriksen}, \bibinfo{author}{C.~L. Brasen}, \bibinfo{author}{O.~Hilberg}, \bibinfo{author}{T.~F. Hansen}, \bibinfo{author}{L.~H. Jensen}, \bibinfo{author}{A.~Peimankar}, \bibinfo{author}{U.~K. Wiil},
\newblock \bibinfo{title}{Identification of patients’ smoking status using an explainable ai approach: a danish electronic health records case study},
\newblock \bibinfo{journal}{BMC Medical Research Methodology} \bibinfo{volume}{24} (\bibinfo{year}{2024}) \bibinfo{pages}{114}.
\bibitem[{Wang et~al.(2019)Wang, Luo, Wang, Wampfler, Yang, and Liu}]{wang2019natural}
\bibinfo{author}{L.~Wang}, \bibinfo{author}{L.~Luo}, \bibinfo{author}{Y.~Wang}, \bibinfo{author}{J.~Wampfler}, \bibinfo{author}{P.~Yang}, \bibinfo{author}{H.~Liu},
\newblock \bibinfo{title}{Natural language processing for populating lung cancer clinical research data},
\newblock \bibinfo{journal}{BMC medical informatics and decision making} \bibinfo{volume}{19} (\bibinfo{year}{2019}) \bibinfo{pages}{1--10}.
\bibitem[{Yu et~al.(2021)Yu, Yang, Dang, Wu, Adekkanattu, Pathak, George, Hogan, Guo, Bian et~al.}]{yu2021study}
\bibinfo{author}{Z.~Yu}, \bibinfo{author}{X.~Yang}, \bibinfo{author}{C.~Dang}, \bibinfo{author}{S.~Wu}, \bibinfo{author}{P.~Adekkanattu}, \bibinfo{author}{J.~Pathak}, \bibinfo{author}{T.~J. George}, \bibinfo{author}{W.~R. Hogan}, \bibinfo{author}{Y.~Guo}, \bibinfo{author}{J.~Bian}, et~al.,
\newblock \bibinfo{title}{A study of social and behavioral determinants of health in lung cancer patients using transformers-based natural language processing models},
\newblock in: \bibinfo{booktitle}{AMIA Annual Symposium Proceedings}, volume \bibinfo{volume}{2021}, \bibinfo{organization}{American Medical Informatics Association}, \bibinfo{year}{2021}, p. \bibinfo{pages}{1225}.
\bibitem[{Zigman~Suchsland et~al.(2022)Zigman~Suchsland, Kowalski, Burkhardt, Prado, Kessler, Yetisgen, Au, Stephens, Farjah, Schleyer et~al.}]{zigman2022timely}
\bibinfo{author}{M.~Zigman~Suchsland}, \bibinfo{author}{L.~Kowalski}, \bibinfo{author}{H.~A. Burkhardt}, \bibinfo{author}{M.~G. Prado}, \bibinfo{author}{L.~G. Kessler}, \bibinfo{author}{M.~Yetisgen}, \bibinfo{author}{M.~A. Au}, \bibinfo{author}{K.~A. Stephens}, \bibinfo{author}{F.~Farjah}, \bibinfo{author}{A.~M. Schleyer}, et~al.,
\newblock \bibinfo{title}{How timely is diagnosis of lung cancer? cohort study of individuals with lung cancer presenting in ambulatory care in the united states},
\newblock \bibinfo{journal}{Cancers} \bibinfo{volume}{14} (\bibinfo{year}{2022}) \bibinfo{pages}{5756}.
\bibitem[{Yu et~al.(2022)Yu, Yang, Guo, Bian, and Wu}]{yu2022assessing}
\bibinfo{author}{Z.~Yu}, \bibinfo{author}{X.~Yang}, \bibinfo{author}{Y.~Guo}, \bibinfo{author}{J.~Bian}, \bibinfo{author}{Y.~Wu},
\newblock \bibinfo{title}{Assessing the documentation of social determinants of health for lung cancer patients in clinical narratives},
\newblock \bibinfo{journal}{Frontiers in public health} \bibinfo{volume}{10} (\bibinfo{year}{2022}) \bibinfo{pages}{778463}.
\bibitem[{Benedum et~al.(2023)Benedum, Sondhi, Fidyk, Cohen, Nemeth, Adamson, Est{\'e}vez, and Bozkurt}]{benedum2023replication}
\bibinfo{author}{C.~M. Benedum}, \bibinfo{author}{A.~Sondhi}, \bibinfo{author}{E.~Fidyk}, \bibinfo{author}{A.~B. Cohen}, \bibinfo{author}{S.~Nemeth}, \bibinfo{author}{B.~Adamson}, \bibinfo{author}{M.~Est{\'e}vez}, \bibinfo{author}{S.~Bozkurt},
\newblock \bibinfo{title}{Replication of real-world evidence in oncology using electronic health record data extracted by machine learning},
\newblock \bibinfo{journal}{Cancers} \bibinfo{volume}{15} (\bibinfo{year}{2023}) \bibinfo{pages}{1853}.
\bibitem[{Hochheiser et~al.(2023)Hochheiser, Finan, Yuan, Durbin, Jeong, Hands, Rust, Kavuluru, Wu, Warner et~al.}]{hochheiser2023deepphe}
\bibinfo{author}{H.~Hochheiser}, \bibinfo{author}{S.~Finan}, \bibinfo{author}{Z.~Yuan}, \bibinfo{author}{E.~B. Durbin}, \bibinfo{author}{J.~C. Jeong}, \bibinfo{author}{I.~Hands}, \bibinfo{author}{D.~Rust}, \bibinfo{author}{R.~Kavuluru}, \bibinfo{author}{X.-C. Wu}, \bibinfo{author}{J.~L. Warner}, et~al.,
\newblock \bibinfo{title}{Deepphe-cr: Natural language processing software services for cancer registrar case abstraction},
\newblock \bibinfo{journal}{JCO Clinical Cancer Informatics} \bibinfo{volume}{7} (\bibinfo{year}{2023}) \bibinfo{pages}{e2300156}.
\bibitem[{Huang et~al.(2024)Huang, Yang, Rong, Nezafati, Treager, Chi, Wang, Cheng, Guo, Klesse et~al.}]{huang2024critical}
\bibinfo{author}{J.~Huang}, \bibinfo{author}{D.~M. Yang}, \bibinfo{author}{R.~Rong}, \bibinfo{author}{K.~Nezafati}, \bibinfo{author}{C.~Treager}, \bibinfo{author}{Z.~Chi}, \bibinfo{author}{S.~Wang}, \bibinfo{author}{X.~Cheng}, \bibinfo{author}{Y.~Guo}, \bibinfo{author}{L.~J. Klesse}, et~al.,
\newblock \bibinfo{title}{A critical assessment of using chatgpt for extracting structured data from clinical notes},
\newblock \bibinfo{journal}{npj Digital Medicine} \bibinfo{volume}{7} (\bibinfo{year}{2024}) \bibinfo{pages}{106}.
\bibitem[{Morin et~al.(2021)Morin, Valli{\`e}res, Braunstein, Ginart, Upadhaya, Woodruff, Zwanenburg, Chatterjee, Villanueva-Meyer, Valdes et~al.}]{morin2021artificial}
\bibinfo{author}{O.~Morin}, \bibinfo{author}{M.~Valli{\`e}res}, \bibinfo{author}{S.~Braunstein}, \bibinfo{author}{J.~B. Ginart}, \bibinfo{author}{T.~Upadhaya}, \bibinfo{author}{H.~C. Woodruff}, \bibinfo{author}{A.~Zwanenburg}, \bibinfo{author}{A.~Chatterjee}, \bibinfo{author}{J.~E. Villanueva-Meyer}, \bibinfo{author}{G.~Valdes}, et~al.,
\newblock \bibinfo{title}{An artificial intelligence framework integrating longitudinal electronic health records with real-world data enables continuous pan-cancer prognostication},
\newblock \bibinfo{journal}{Nature Cancer} \bibinfo{volume}{2} (\bibinfo{year}{2021}) \bibinfo{pages}{709--722}.
\bibitem[{Kehl et~al.(2021)Kehl, Groha, Lepisto, Elmarakeby, Lindsay, Gusev, Van~Allen, Hassett, and Schrag}]{kehl2021clinical}
\bibinfo{author}{K.~L. Kehl}, \bibinfo{author}{S.~Groha}, \bibinfo{author}{E.~M. Lepisto}, \bibinfo{author}{H.~Elmarakeby}, \bibinfo{author}{J.~Lindsay}, \bibinfo{author}{A.~Gusev}, \bibinfo{author}{E.~M. Van~Allen}, \bibinfo{author}{M.~J. Hassett}, \bibinfo{author}{D.~Schrag},
\newblock \bibinfo{title}{Clinical inflection point detection on the basis of ehr data to identify clinical trial--ready patients with cancer},
\newblock \bibinfo{journal}{JCO Clinical Cancer Informatics} \bibinfo{volume}{5} (\bibinfo{year}{2021}) \bibinfo{pages}{622--630}.
\bibitem[{Bhatt et~al.(2023)Bhatt, Johnson, Markovitz, Gray, Nipp, Ufere, Rice, Reynolds, Lavoie, Clay et~al.}]{bhatt2023use}
\bibinfo{author}{S.~Bhatt}, \bibinfo{author}{P.~C. Johnson}, \bibinfo{author}{N.~H. Markovitz}, \bibinfo{author}{T.~Gray}, \bibinfo{author}{R.~D. Nipp}, \bibinfo{author}{N.~Ufere}, \bibinfo{author}{J.~Rice}, \bibinfo{author}{M.~J. Reynolds}, \bibinfo{author}{M.~W. Lavoie}, \bibinfo{author}{M.~A. Clay}, et~al.,
\newblock \bibinfo{title}{The use of natural language processing to assess social support in patients with advanced cancer},
\newblock \bibinfo{journal}{The Oncologist} \bibinfo{volume}{28} (\bibinfo{year}{2023}) \bibinfo{pages}{165--171}.
\bibitem[{Kehl et~al.(2021)Kehl, Xu, Gusev, Bakouny, Choueiri, Riaz, Elmarakeby, Van~Allen, and Schrag}]{kehl2021artificial}
\bibinfo{author}{K.~L. Kehl}, \bibinfo{author}{W.~Xu}, \bibinfo{author}{A.~Gusev}, \bibinfo{author}{Z.~Bakouny}, \bibinfo{author}{T.~K. Choueiri}, \bibinfo{author}{I.~B. Riaz}, \bibinfo{author}{H.~Elmarakeby}, \bibinfo{author}{E.~M. Van~Allen}, \bibinfo{author}{D.~Schrag},
\newblock \bibinfo{title}{Artificial intelligence-aided clinical annotation of a large multi-cancer genomic dataset},
\newblock \bibinfo{journal}{Nature communications} \bibinfo{volume}{12} (\bibinfo{year}{2021}) \bibinfo{pages}{7304}.
\bibitem[{Yu et~al.(2024)Yu, Peng, Yang, Dang, Adekkanattu, Patra, Peng, Pathak, Wilson, Chang et~al.}]{yu2024identifying}
\bibinfo{author}{Z.~Yu}, \bibinfo{author}{C.~Peng}, \bibinfo{author}{X.~Yang}, \bibinfo{author}{C.~Dang}, \bibinfo{author}{P.~Adekkanattu}, \bibinfo{author}{B.~G. Patra}, \bibinfo{author}{Y.~Peng}, \bibinfo{author}{J.~Pathak}, \bibinfo{author}{D.~L. Wilson}, \bibinfo{author}{C.-Y. Chang}, et~al.,
\newblock \bibinfo{title}{Identifying social determinants of health from clinical narratives: A study of performance, documentation ratio, and potential bias},
\newblock \bibinfo{journal}{Journal of biomedical informatics} \bibinfo{volume}{153} (\bibinfo{year}{2024}) \bibinfo{pages}{104642}.
\bibitem[{Araki et~al.(2023)Araki, Matsumoto, Togo, Yonemoto, Ohki, Xu, Hasegawa, Inoue, Yamashita, and Miyazaki}]{araki2023real}
\bibinfo{author}{K.~Araki}, \bibinfo{author}{N.~Matsumoto}, \bibinfo{author}{K.~Togo}, \bibinfo{author}{N.~Yonemoto}, \bibinfo{author}{E.~Ohki}, \bibinfo{author}{L.~Xu}, \bibinfo{author}{Y.~Hasegawa}, \bibinfo{author}{H.~Inoue}, \bibinfo{author}{S.~Yamashita}, \bibinfo{author}{T.~Miyazaki},
\newblock \bibinfo{title}{Real-world treatment response in japanese patients with cancer using unstructured data from electronic health records},
\newblock \bibinfo{journal}{Health and Technology} \bibinfo{volume}{13} (\bibinfo{year}{2023}) \bibinfo{pages}{253--262}.
\bibitem[{Wang et~al.(2019)Wang, Wampfler, Dispenzieri, Xu, Yang, and Liu}]{wang2019achievability}
\bibinfo{author}{L.~Wang}, \bibinfo{author}{J.~Wampfler}, \bibinfo{author}{A.~Dispenzieri}, \bibinfo{author}{H.~Xu}, \bibinfo{author}{P.~Yang}, \bibinfo{author}{H.~Liu},
\newblock \bibinfo{title}{Achievability to extract specific date information for cancer research},
\newblock in: \bibinfo{booktitle}{AMIA Annual Symposium Proceedings}, volume \bibinfo{volume}{2019}, \bibinfo{organization}{American Medical Informatics Association}, \bibinfo{year}{2019}, p. \bibinfo{pages}{893}.
\bibitem[{Paul et~al.(2022)Paul, Rana, Tautam, Kotapati, Jampani, Singh, Islam, Mandhadi, Sharma, Barnes et~al.}]{paul2022investigation}
\bibinfo{author}{T.~Paul}, \bibinfo{author}{M.~K.~Z. Rana}, \bibinfo{author}{P.~A. Tautam}, \bibinfo{author}{T.~V.~P. Kotapati}, \bibinfo{author}{Y.~Jampani}, \bibinfo{author}{N.~Singh}, \bibinfo{author}{H.~Islam}, \bibinfo{author}{V.~Mandhadi}, \bibinfo{author}{V.~Sharma}, \bibinfo{author}{M.~Barnes}, et~al.,
\newblock \bibinfo{title}{Investigation of the utility of features in a clinical de-identification model: A demonstration using ehr pathology reports for advanced nsclc patients},
\newblock \bibinfo{journal}{Frontiers in digital health} \bibinfo{volume}{4} (\bibinfo{year}{2022}) \bibinfo{pages}{728922}.
\bibitem[{Kersloot et~al.(2019)Kersloot, Lau, Abu-Hanna, Arts, and Cornet}]{kersloot2019automated}
\bibinfo{author}{M.~G. Kersloot}, \bibinfo{author}{F.~Lau}, \bibinfo{author}{A.~Abu-Hanna}, \bibinfo{author}{D.~L. Arts}, \bibinfo{author}{R.~Cornet},
\newblock \bibinfo{title}{Automated snomed ct concept and attribute relationship detection through a web-based implementation of ctakes},
\newblock \bibinfo{journal}{Journal of biomedical semantics} \bibinfo{volume}{10} (\bibinfo{year}{2019}) \bibinfo{pages}{1--13}.
\bibitem[{Adamson et~al.(2023)Adamson, Waskom, Blarre, Kelly, Krismer, Nemeth, Gippetti, Ritten, Harrison, Ho et~al.}]{adamson2023approach}
\bibinfo{author}{B.~Adamson}, \bibinfo{author}{M.~Waskom}, \bibinfo{author}{A.~Blarre}, \bibinfo{author}{J.~Kelly}, \bibinfo{author}{K.~Krismer}, \bibinfo{author}{S.~Nemeth}, \bibinfo{author}{J.~Gippetti}, \bibinfo{author}{J.~Ritten}, \bibinfo{author}{K.~Harrison}, \bibinfo{author}{G.~Ho}, et~al.,
\newblock \bibinfo{title}{Approach to machine learning for extraction of real-world data variables from electronic health records},
\newblock \bibinfo{journal}{Frontiers in Pharmacology} \bibinfo{volume}{14} (\bibinfo{year}{2023}) \bibinfo{pages}{1180962}.
\bibitem[{Yusuf et~al.(2024)Yusuf, Boyne, O’Sullivan, Brenner, Cheung, Mirza, and Jarada}]{yusuf2024text}
\bibinfo{author}{A.~Yusuf}, \bibinfo{author}{D.~J. Boyne}, \bibinfo{author}{D.~E. O’Sullivan}, \bibinfo{author}{D.~R. Brenner}, \bibinfo{author}{W.~Y. Cheung}, \bibinfo{author}{I.~Mirza}, \bibinfo{author}{T.~N. Jarada},
\newblock \bibinfo{title}{Text analysis framework for identifying mutations among non-small cell lung cancer patients from laboratory data},
\newblock \bibinfo{journal}{BMC Medical Research Methodology} \bibinfo{volume}{24} (\bibinfo{year}{2024}) \bibinfo{pages}{63}.
\bibitem[{Seesaghur et~al.(2023)Seesaghur, Egger, Warden, Abbasi, Levick, Riaz, McMahon, Thompson, and Cheeseman}]{seesaghur2023assessment}
\bibinfo{author}{A.~Seesaghur}, \bibinfo{author}{P.~Egger}, \bibinfo{author}{J.~Warden}, \bibinfo{author}{A.~Abbasi}, \bibinfo{author}{B.~Levick}, \bibinfo{author}{M.~Riaz}, \bibinfo{author}{P.~McMahon}, \bibinfo{author}{M.~Thompson}, \bibinfo{author}{S.~Cheeseman},
\newblock \bibinfo{title}{Assessment of bone-targeting agents use in patients with bone metastasis from breast, lung or prostate cancer using structured and unstructured electronic health records from a regional uk-based hospital},
\newblock \bibinfo{journal}{BMJ open} \bibinfo{volume}{13} (\bibinfo{year}{2023}) \bibinfo{pages}{e069214}.
\bibitem[{Li et~al.(2023)Li, Wang, Cai, Sun, Yang, Liu, Wang, and Lv}]{li2023interpretable}
\bibinfo{author}{J.~Li}, \bibinfo{author}{X.~Wang}, \bibinfo{author}{L.~Cai}, \bibinfo{author}{J.~Sun}, \bibinfo{author}{Z.~Yang}, \bibinfo{author}{W.~Liu}, \bibinfo{author}{Z.~Wang}, \bibinfo{author}{H.~Lv},
\newblock \bibinfo{title}{An interpretable deep learning framework for predicting liver metastases in postoperative colorectal cancer patients using natural language processing and clinical data integration},
\newblock \bibinfo{journal}{Cancer Medicine} \bibinfo{volume}{12} (\bibinfo{year}{2023}) \bibinfo{pages}{19337--19351}.
\bibitem[{Laique et~al.(2021)Laique, Hayat, Sarvepalli, Vaughn, Ibrahim, McMichael, Qaiser, Burke, Bhatt, Rhodes et~al.}]{laique2021application}
\bibinfo{author}{S.~N. Laique}, \bibinfo{author}{U.~Hayat}, \bibinfo{author}{S.~Sarvepalli}, \bibinfo{author}{B.~Vaughn}, \bibinfo{author}{M.~Ibrahim}, \bibinfo{author}{J.~McMichael}, \bibinfo{author}{K.~N. Qaiser}, \bibinfo{author}{C.~Burke}, \bibinfo{author}{A.~Bhatt}, \bibinfo{author}{C.~Rhodes}, et~al.,
\newblock \bibinfo{title}{Application of optical character recognition with natural language processing for large-scale quality metric data extraction in colonoscopy reports},
\newblock \bibinfo{journal}{Gastrointestinal endoscopy} \bibinfo{volume}{93} (\bibinfo{year}{2021}) \bibinfo{pages}{750--757}.
\bibitem[{Agaronnik et~al.(2020)Agaronnik, Lindvall, El-Jawahri, He, and Iezzoni}]{agaronnik2020challenges}
\bibinfo{author}{N.~D. Agaronnik}, \bibinfo{author}{C.~Lindvall}, \bibinfo{author}{A.~El-Jawahri}, \bibinfo{author}{W.~He}, \bibinfo{author}{L.~I. Iezzoni},
\newblock \bibinfo{title}{Challenges of developing a natural language processing method with electronic health records to identify persons with chronic mobility disability},
\newblock \bibinfo{journal}{Archives of physical medicine and rehabilitation} \bibinfo{volume}{101} (\bibinfo{year}{2020}) \bibinfo{pages}{1739--1746}.
\bibitem[{Tamm et~al.(2022)Tamm, Jones, Perry, Campbell, Carten, Davies, Galdikas, English, Garbett, Glampson et~al.}]{tamm2022establishing}
\bibinfo{author}{A.~Tamm}, \bibinfo{author}{H.~J. Jones}, \bibinfo{author}{W.~Perry}, \bibinfo{author}{D.~Campbell}, \bibinfo{author}{R.~Carten}, \bibinfo{author}{J.~Davies}, \bibinfo{author}{A.~Galdikas}, \bibinfo{author}{L.~English}, \bibinfo{author}{A.~Garbett}, \bibinfo{author}{B.~Glampson}, et~al.,
\newblock \bibinfo{title}{Establishing a colorectal cancer research database from routinely collected health data: the process and potential from a pilot study},
\newblock \bibinfo{journal}{BMJ Health \& Care Informatics} \bibinfo{volume}{29} (\bibinfo{year}{2022}).
\bibitem[{Karwa et~al.(2020)Karwa, Patell, Parthasarathy, Lopez, McMichael, and Burke}]{karwa2020development}
\bibinfo{author}{A.~Karwa}, \bibinfo{author}{R.~Patell}, \bibinfo{author}{G.~Parthasarathy}, \bibinfo{author}{R.~Lopez}, \bibinfo{author}{J.~McMichael}, \bibinfo{author}{C.~A. Burke},
\newblock \bibinfo{title}{Development of an automated algorithm to generate guideline-based recommendations for follow-up colonoscopy},
\newblock \bibinfo{journal}{Clinical Gastroenterology and Hepatology} \bibinfo{volume}{18} (\bibinfo{year}{2020}) \bibinfo{pages}{2038--2045}.
\bibitem[{Luo et~al.(2021)Luo, Gandhi, Storey, Zhang, Han, and Huang}]{luo2021computational}
\bibinfo{author}{X.~Luo}, \bibinfo{author}{P.~Gandhi}, \bibinfo{author}{S.~Storey}, \bibinfo{author}{Z.~Zhang}, \bibinfo{author}{Z.~Han}, \bibinfo{author}{K.~Huang},
\newblock \bibinfo{title}{A computational framework to analyze the associations between symptoms and cancer patient attributes post chemotherapy using ehr data},
\newblock \bibinfo{journal}{IEEE Journal of Biomedical and Health Informatics} \bibinfo{volume}{25} (\bibinfo{year}{2021}) \bibinfo{pages}{4098--4109}.
\bibitem[{Do et~al.(2021)Do, Lupton, Causa~Andrieu, Luthra, Taya, Batch, Nguyen, Rahurkar, Gazit, Nicholas et~al.}]{do2021patterns}
\bibinfo{author}{R.~K. Do}, \bibinfo{author}{K.~Lupton}, \bibinfo{author}{P.~I. Causa~Andrieu}, \bibinfo{author}{A.~Luthra}, \bibinfo{author}{M.~Taya}, \bibinfo{author}{K.~Batch}, \bibinfo{author}{H.~Nguyen}, \bibinfo{author}{P.~Rahurkar}, \bibinfo{author}{L.~Gazit}, \bibinfo{author}{K.~Nicholas}, et~al.,
\newblock \bibinfo{title}{Patterns of metastatic disease in patients with cancer derived from natural language processing of structured ct radiology reports over a 10-year period},
\newblock \bibinfo{journal}{Radiology} \bibinfo{volume}{301} (\bibinfo{year}{2021}) \bibinfo{pages}{115--122}.
\bibitem[{Luo et~al.(2021)Luo, Storey, Gandhi, Zhang, Metzger, and Huang}]{luo2021analyzing}
\bibinfo{author}{X.~Luo}, \bibinfo{author}{S.~Storey}, \bibinfo{author}{P.~Gandhi}, \bibinfo{author}{Z.~Zhang}, \bibinfo{author}{M.~Metzger}, \bibinfo{author}{K.~Huang},
\newblock \bibinfo{title}{Analyzing the symptoms in colorectal and breast cancer patients with or without type 2 diabetes using ehr data},
\newblock \bibinfo{journal}{Health Informatics Journal} \bibinfo{volume}{27} (\bibinfo{year}{2021}) \bibinfo{pages}{14604582211000785}.
\bibitem[{Shi et~al.(2022)Shi, Morgan, Bradshaw, Jung, Kohlmann, Kaphingst, Kawamoto, Del~Fiol et~al.}]{shi2022identifying}
\bibinfo{author}{J.~Shi}, \bibinfo{author}{K.~L. Morgan}, \bibinfo{author}{R.~L. Bradshaw}, \bibinfo{author}{S.-H. Jung}, \bibinfo{author}{W.~Kohlmann}, \bibinfo{author}{K.~A. Kaphingst}, \bibinfo{author}{K.~Kawamoto}, \bibinfo{author}{G.~Del~Fiol}, et~al.,
\newblock \bibinfo{title}{Identifying patients who meet criteria for genetic testing of hereditary cancers based on structured and unstructured family health history data in the electronic health record: natural language processing approach},
\newblock \bibinfo{journal}{JMIR Medical Informatics} \bibinfo{volume}{10} (\bibinfo{year}{2022}) \bibinfo{pages}{e37842}.
\bibitem[{Ryu et~al.(2020)Ryu, Yoon, Kim, Lee, Baek, Yi, Na, Kim, Baek, Hwang et~al.}]{ryu2020transformation}
\bibinfo{author}{B.~Ryu}, \bibinfo{author}{E.~Yoon}, \bibinfo{author}{S.~Kim}, \bibinfo{author}{S.~Lee}, \bibinfo{author}{H.~Baek}, \bibinfo{author}{S.~Yi}, \bibinfo{author}{H.~Y. Na}, \bibinfo{author}{J.-W. Kim}, \bibinfo{author}{R.-M. Baek}, \bibinfo{author}{H.~Hwang}, et~al.,
\newblock \bibinfo{title}{Transformation of pathology reports into the common data model with oncology module: use case for colon cancer},
\newblock \bibinfo{journal}{Journal of medical Internet research} \bibinfo{volume}{22} (\bibinfo{year}{2020}) \bibinfo{pages}{e18526}.
\bibitem[{Sarwal et~al.(2024)Sarwal, Wang, Gandhi, Pour, Janssens, Delgado, Doering, Mishra, Greenwood, Liu et~al.}]{sarwal2024identification}
\bibinfo{author}{D.~Sarwal}, \bibinfo{author}{L.~Wang}, \bibinfo{author}{S.~Gandhi}, \bibinfo{author}{E.~S.~H. Pour}, \bibinfo{author}{L.~P. Janssens}, \bibinfo{author}{A.~M. Delgado}, \bibinfo{author}{K.~A. Doering}, \bibinfo{author}{A.~K. Mishra}, \bibinfo{author}{J.~D. Greenwood}, \bibinfo{author}{H.~Liu}, et~al.,
\newblock \bibinfo{title}{Identification of pancreatic cancer risk factors from clinical notes using natural language processing},
\newblock \bibinfo{journal}{Pancreatology}  (\bibinfo{year}{2024}).
\bibitem[{Harrison et~al.(2023)Harrison, Yala, Mikhael, Roldan, Ciprani, Michelakos, Bolm, Qadan, Ferrone, Fernandez-del Castillo et~al.}]{harrison2023successful}
\bibinfo{author}{J.~M. Harrison}, \bibinfo{author}{A.~Yala}, \bibinfo{author}{P.~Mikhael}, \bibinfo{author}{J.~Roldan}, \bibinfo{author}{D.~Ciprani}, \bibinfo{author}{T.~Michelakos}, \bibinfo{author}{L.~Bolm}, \bibinfo{author}{M.~Qadan}, \bibinfo{author}{C.~Ferrone}, \bibinfo{author}{C.~Fernandez-del Castillo}, et~al.,
\newblock \bibinfo{title}{Successful development of a natural language processing algorithm for pancreatic neoplasms and associated histologic features},
\newblock \bibinfo{journal}{Pancreas} \bibinfo{volume}{52} (\bibinfo{year}{2023}) \bibinfo{pages}{e219--e223}.
\bibitem[{Chen et~al.(2023)Chen, Guevara, Ramirez, Murray, Warner, Aerts, Miller, Savova, Mak, and Bitterman}]{chen2023natural}
\bibinfo{author}{S.~Chen}, \bibinfo{author}{M.~Guevara}, \bibinfo{author}{N.~Ramirez}, \bibinfo{author}{A.~Murray}, \bibinfo{author}{J.~L. Warner}, \bibinfo{author}{H.~J. Aerts}, \bibinfo{author}{T.~A. Miller}, \bibinfo{author}{G.~K. Savova}, \bibinfo{author}{R.~H. Mak}, \bibinfo{author}{D.~S. Bitterman},
\newblock \bibinfo{title}{Natural language processing to automatically extract the presence and severity of esophagitis in notes of patients undergoing radiotherapy},
\newblock \bibinfo{journal}{JCO Clinical Cancer Informatics} \bibinfo{volume}{7} (\bibinfo{year}{2023}) \bibinfo{pages}{e2300048}.
\bibitem[{Karimi et~al.(2021)Karimi, Blayney, Kurian, Shen, Yamashita, Rubin, and Banerjee}]{karimi2021development}
\bibinfo{author}{Y.~H. Karimi}, \bibinfo{author}{D.~W. Blayney}, \bibinfo{author}{A.~W. Kurian}, \bibinfo{author}{J.~Shen}, \bibinfo{author}{R.~Yamashita}, \bibinfo{author}{D.~Rubin}, \bibinfo{author}{I.~Banerjee},
\newblock \bibinfo{title}{Development and use of natural language processing for identification of distant cancer recurrence and sites of distant recurrence using unstructured electronic health record data},
\newblock \bibinfo{journal}{JCO Clinical Cancer Informatics} \bibinfo{volume}{5} (\bibinfo{year}{2021}) \bibinfo{pages}{469--478}.
\bibitem[{Zhou et~al.(2023)Zhou, Wang, Wang, Sun, Blaes, Liu, and Zhang}]{zhou2023cross}
\bibinfo{author}{S.~Zhou}, \bibinfo{author}{N.~Wang}, \bibinfo{author}{L.~Wang}, \bibinfo{author}{J.~Sun}, \bibinfo{author}{A.~Blaes}, \bibinfo{author}{H.~Liu}, \bibinfo{author}{R.~Zhang},
\newblock \bibinfo{title}{A cross-institutional evaluation on breast cancer phenotyping nlp algorithms on electronic health records},
\newblock \bibinfo{journal}{Computational and Structural Biotechnology Journal} \bibinfo{volume}{22} (\bibinfo{year}{2023}) \bibinfo{pages}{32--40}.
\bibitem[{Zhou et~al.(2022)Zhou, Wang, Wang, Liu, and Zhang}]{zhou2022cancerbert}
\bibinfo{author}{S.~Zhou}, \bibinfo{author}{N.~Wang}, \bibinfo{author}{L.~Wang}, \bibinfo{author}{H.~Liu}, \bibinfo{author}{R.~Zhang},
\newblock \bibinfo{title}{Cancerbert: a cancer domain-specific language model for extracting breast cancer phenotypes from electronic health records},
\newblock \bibinfo{journal}{Journal of the American Medical Informatics Association} \bibinfo{volume}{29} (\bibinfo{year}{2022}) \bibinfo{pages}{1208--1216}.
\bibitem[{Garc{\'\i}a-Barrag{\'a}n et~al.(2024)Garc{\'\i}a-Barrag{\'a}n, Gonz{\'a}lez~Calatayud, Solarte-Pab{\'o}n, Provencio, Menasalvas, and Robles}]{garcia2024gpt}
\bibinfo{author}{{\'A}.~Garc{\'\i}a-Barrag{\'a}n}, \bibinfo{author}{A.~Gonz{\'a}lez~Calatayud}, \bibinfo{author}{O.~Solarte-Pab{\'o}n}, \bibinfo{author}{M.~Provencio}, \bibinfo{author}{E.~Menasalvas}, \bibinfo{author}{V.~Robles},
\newblock \bibinfo{title}{Gpt for medical entity recognition in spanish},
\newblock \bibinfo{journal}{Multimedia Tools and Applications}  (\bibinfo{year}{2024}) \bibinfo{pages}{1--20}.
\bibitem[{Ribelles et~al.(2021)Ribelles, Jerez, Rodriguez-Brazzarola, Jimenez, Diaz-Redondo, Mesa, Marquez, Sanchez-Mu{\~n}oz, Pajares, Carabantes et~al.}]{ribelles2021machine}
\bibinfo{author}{N.~Ribelles}, \bibinfo{author}{J.~M. Jerez}, \bibinfo{author}{P.~Rodriguez-Brazzarola}, \bibinfo{author}{B.~Jimenez}, \bibinfo{author}{T.~Diaz-Redondo}, \bibinfo{author}{H.~Mesa}, \bibinfo{author}{A.~Marquez}, \bibinfo{author}{A.~Sanchez-Mu{\~n}oz}, \bibinfo{author}{B.~Pajares}, \bibinfo{author}{F.~Carabantes}, et~al.,
\newblock \bibinfo{title}{Machine learning and natural language processing (nlp) approach to predict early progression to first-line treatment in real-world hormone receptor-positive (hr+)/her2-negative advanced breast cancer patients},
\newblock \bibinfo{journal}{European Journal of Cancer} \bibinfo{volume}{144} (\bibinfo{year}{2021}) \bibinfo{pages}{224--231}.
\bibitem[{Sanyal et~al.(2021)Sanyal, Tariq, Kurian, Rubin, and Banerjee}]{sanyal2021weakly}
\bibinfo{author}{J.~Sanyal}, \bibinfo{author}{A.~Tariq}, \bibinfo{author}{A.~W. Kurian}, \bibinfo{author}{D.~Rubin}, \bibinfo{author}{I.~Banerjee},
\newblock \bibinfo{title}{Weakly supervised temporal model for prediction of breast cancer distant recurrence},
\newblock \bibinfo{journal}{Scientific reports} \bibinfo{volume}{11} (\bibinfo{year}{2021}) \bibinfo{pages}{9461}.
\bibitem[{Thompson et~al.(2019)Thompson, Hu, Mudaranthakam, Streeter, Neums, Park, Koestler, Gajewski, Jensen, and Mayo}]{thompson2019relevant}
\bibinfo{author}{J.~Thompson}, \bibinfo{author}{J.~Hu}, \bibinfo{author}{D.~P. Mudaranthakam}, \bibinfo{author}{D.~Streeter}, \bibinfo{author}{L.~Neums}, \bibinfo{author}{M.~Park}, \bibinfo{author}{D.~C. Koestler}, \bibinfo{author}{B.~Gajewski}, \bibinfo{author}{R.~Jensen}, \bibinfo{author}{M.~S. Mayo},
\newblock \bibinfo{title}{Relevant word order vectorization for improved natural language processing in electronic health records},
\newblock \bibinfo{journal}{Scientific reports} \bibinfo{volume}{9} (\bibinfo{year}{2019}) \bibinfo{pages}{9253}.
\bibitem[{Zeng et~al.(2019)Zeng, Yao, Roy, Li, Espino, Clare, Khan, and Luo}]{zeng2019identifying}
\bibinfo{author}{Z.~Zeng}, \bibinfo{author}{L.~Yao}, \bibinfo{author}{A.~Roy}, \bibinfo{author}{X.~Li}, \bibinfo{author}{S.~Espino}, \bibinfo{author}{S.~E. Clare}, \bibinfo{author}{S.~A. Khan}, \bibinfo{author}{Y.~Luo},
\newblock \bibinfo{title}{Identifying breast cancer distant recurrences from electronic health records using machine learning},
\newblock \bibinfo{journal}{Journal of healthcare informatics research} \bibinfo{volume}{3} (\bibinfo{year}{2019}) \bibinfo{pages}{283--299}.
\bibitem[{Jin et~al.(2021)Jin, Junren, Jingwen, Yajing, Xi, and Ke}]{jin2021research}
\bibinfo{author}{Y.~Jin}, \bibinfo{author}{W.~Junren}, \bibinfo{author}{J.~Jingwen}, \bibinfo{author}{S.~Yajing}, \bibinfo{author}{C.~Xi}, \bibinfo{author}{Q.~Ke},
\newblock \bibinfo{title}{Research on the construction and application of breast cancer-specific database system based on full data lifecycle},
\newblock \bibinfo{journal}{Frontiers in Public Health} \bibinfo{volume}{9} (\bibinfo{year}{2021}) \bibinfo{pages}{712827}.
\bibitem[{Chen et~al.(2022)Chen, Hao, Zou, Hollander, Ng, and Isaac}]{chen2022automated}
\bibinfo{author}{Y.~Chen}, \bibinfo{author}{L.~Hao}, \bibinfo{author}{V.~Z. Zou}, \bibinfo{author}{Z.~Hollander}, \bibinfo{author}{R.~T. Ng}, \bibinfo{author}{K.~V. Isaac},
\newblock \bibinfo{title}{Automated medical chart review for breast cancer outcomes research: a novel natural language processing extraction system},
\newblock \bibinfo{journal}{BMC medical research methodology} \bibinfo{volume}{22} (\bibinfo{year}{2022}) \bibinfo{pages}{136}.
\bibitem[{Brizzi et~al.(2020)Brizzi, Zupanc, Udelsman, Tulsky, Wright, Poort, and Lindvall}]{brizzi2020natural}
\bibinfo{author}{K.~Brizzi}, \bibinfo{author}{S.~N. Zupanc}, \bibinfo{author}{B.~V. Udelsman}, \bibinfo{author}{J.~A. Tulsky}, \bibinfo{author}{A.~A. Wright}, \bibinfo{author}{H.~Poort}, \bibinfo{author}{C.~Lindvall},
\newblock \bibinfo{title}{Natural language processing to assess palliative care and end-of-life process measures in patients with breast cancer with leptomeningeal disease},
\newblock \bibinfo{journal}{American Journal of Hospice and Palliative Medicine{\textregistered}} \bibinfo{volume}{37} (\bibinfo{year}{2020}) \bibinfo{pages}{371--376}.
\bibitem[{Alkaitis et~al.(2021)Alkaitis, Agrawal, Riely, Razavi, and Sontag}]{alkaitis2021automated}
\bibinfo{author}{M.~S. Alkaitis}, \bibinfo{author}{M.~N. Agrawal}, \bibinfo{author}{G.~J. Riely}, \bibinfo{author}{P.~Razavi}, \bibinfo{author}{D.~Sontag},
\newblock \bibinfo{title}{Automated nlp extraction of clinical rationale for treatment discontinuation in breast cancer},
\newblock \bibinfo{journal}{JCO Clinical Cancer Informatics} \bibinfo{volume}{5} (\bibinfo{year}{2021}) \bibinfo{pages}{550--560}.
\bibitem[{Solarte-Pab{\'o}n et~al.(2023)Solarte-Pab{\'o}n, Montenegro, Garc{\'\i}a-Barrag{\'a}n, Torrente, Provencio, Menasalvas, and Robles}]{solarte2023transformers}
\bibinfo{author}{O.~Solarte-Pab{\'o}n}, \bibinfo{author}{O.~Montenegro}, \bibinfo{author}{A.~Garc{\'\i}a-Barrag{\'a}n}, \bibinfo{author}{M.~Torrente}, \bibinfo{author}{M.~Provencio}, \bibinfo{author}{E.~Menasalvas}, \bibinfo{author}{V.~Robles},
\newblock \bibinfo{title}{Transformers for extracting breast cancer information from spanish clinical narratives},
\newblock \bibinfo{journal}{Artificial Intelligence in Medicine} \bibinfo{volume}{143} (\bibinfo{year}{2023}) \bibinfo{pages}{102625}.
\bibitem[{Zelina et~al.(2023)Zelina, Hal{\'a}mkov{\'a}, and Nov{\'a}{\v{c}}ek}]{zelina2023extraction}
\bibinfo{author}{P.~Zelina}, \bibinfo{author}{J.~Hal{\'a}mkov{\'a}}, \bibinfo{author}{V.~Nov{\'a}{\v{c}}ek},
\newblock \bibinfo{title}{Extraction, labeling, clustering, and semantic mapping of segments from clinical notes},
\newblock \bibinfo{journal}{IEEE Transactions on NanoBioscience} \bibinfo{volume}{22} (\bibinfo{year}{2023}) \bibinfo{pages}{781--788}.
\bibitem[{Trivedi et~al.(2019)Trivedi, Panahiazar, Liang, Lituiev, Chang, Sohn, Chen, Franc, Joe, and Hadley}]{trivedi2019large}
\bibinfo{author}{H.~M. Trivedi}, \bibinfo{author}{M.~Panahiazar}, \bibinfo{author}{A.~Liang}, \bibinfo{author}{D.~Lituiev}, \bibinfo{author}{P.~Chang}, \bibinfo{author}{J.~H. Sohn}, \bibinfo{author}{Y.-Y. Chen}, \bibinfo{author}{B.~L. Franc}, \bibinfo{author}{B.~Joe}, \bibinfo{author}{D.~Hadley},
\newblock \bibinfo{title}{Large scale semi-automated labeling of routine free-text clinical records for deep learning},
\newblock \bibinfo{journal}{Journal of digital imaging} \bibinfo{volume}{32} (\bibinfo{year}{2019}) \bibinfo{pages}{30--37}.
\bibitem[{Levine et~al.(2019)Levine, Alexander, Sathiyapalan, Agrawal, and Pond}]{levine2019learning}
\bibinfo{author}{M.~N. Levine}, \bibinfo{author}{G.~Alexander}, \bibinfo{author}{A.~Sathiyapalan}, \bibinfo{author}{A.~Agrawal}, \bibinfo{author}{G.~Pond},
\newblock \bibinfo{title}{Learning health system for breast cancer: pilot project experience},
\newblock \bibinfo{journal}{JCO clinical cancer informatics} \bibinfo{volume}{3} (\bibinfo{year}{2019}) \bibinfo{pages}{1--11}.
\bibitem[{Meystre et~al.(2019)Meystre, Heider, Kim, Aruch, and Britten}]{meystre2019automatic}
\bibinfo{author}{S.~M. Meystre}, \bibinfo{author}{P.~M. Heider}, \bibinfo{author}{Y.~Kim}, \bibinfo{author}{D.~B. Aruch}, \bibinfo{author}{C.~D. Britten},
\newblock \bibinfo{title}{Automatic trial eligibility surveillance based on unstructured clinical data},
\newblock \bibinfo{journal}{International journal of medical informatics} \bibinfo{volume}{129} (\bibinfo{year}{2019}) \bibinfo{pages}{13--19}.
\bibitem[{Santus et~al.(2019)Santus, Li, Yala, Peck, Soomro, Faridi, Mamshad, Tang, Lanahan, Barzilay et~al.}]{santus2019neural}
\bibinfo{author}{E.~Santus}, \bibinfo{author}{C.~Li}, \bibinfo{author}{A.~Yala}, \bibinfo{author}{D.~Peck}, \bibinfo{author}{R.~Soomro}, \bibinfo{author}{N.~Faridi}, \bibinfo{author}{I.~Mamshad}, \bibinfo{author}{R.~Tang}, \bibinfo{author}{C.~R. Lanahan}, \bibinfo{author}{R.~Barzilay}, et~al.,
\newblock \bibinfo{title}{Do neural information extraction algorithms generalize across institutions?},
\newblock \bibinfo{journal}{JCO clinical cancer informatics} \bibinfo{volume}{3} (\bibinfo{year}{2019}) \bibinfo{pages}{1--8}.
\bibitem[{Zhao et~al.(2021)Zhao, Weroha, Goode, Liu, and Wang}]{zhao2021generating}
\bibinfo{author}{Y.~Zhao}, \bibinfo{author}{S.~J. Weroha}, \bibinfo{author}{E.~L. Goode}, \bibinfo{author}{H.~Liu}, \bibinfo{author}{C.~Wang},
\newblock \bibinfo{title}{Generating real-world evidence from unstructured clinical notes to examine clinical utility of genetic tests: use case in brcaness},
\newblock \bibinfo{journal}{BMC Medical Informatics and Decision Making} \bibinfo{volume}{21} (\bibinfo{year}{2021}) \bibinfo{pages}{1--13}.
\bibitem[{Yang et~al.(2022)Yang, Zhu, Howard, De~Hoedt, Williams, Freedland, and Klaassen}]{yang2022identification}
\bibinfo{author}{R.~Yang}, \bibinfo{author}{D.~Zhu}, \bibinfo{author}{L.~E. Howard}, \bibinfo{author}{A.~De~Hoedt}, \bibinfo{author}{S.~B. Williams}, \bibinfo{author}{S.~J. Freedland}, \bibinfo{author}{Z.~Klaassen},
\newblock \bibinfo{title}{Identification of patients with metastatic prostate cancer with natural language processing and machine learning},
\newblock \bibinfo{journal}{JCO clinical cancer informatics} \bibinfo{volume}{6} (\bibinfo{year}{2022}) \bibinfo{pages}{e2100071}.
\bibitem[{Alba et~al.(2021)Alba, Gao, Lee, Anglin-Foote, Robison, Katsoulakis, Rose, Efimova, Ferraro, Patterson et~al.}]{alba2021ascertainment}
\bibinfo{author}{P.~R. Alba}, \bibinfo{author}{A.~Gao}, \bibinfo{author}{K.~M. Lee}, \bibinfo{author}{T.~Anglin-Foote}, \bibinfo{author}{B.~Robison}, \bibinfo{author}{E.~Katsoulakis}, \bibinfo{author}{B.~S. Rose}, \bibinfo{author}{O.~Efimova}, \bibinfo{author}{J.~P. Ferraro}, \bibinfo{author}{O.~V. Patterson}, et~al.,
\newblock \bibinfo{title}{Ascertainment of veterans with metastatic prostate cancer in electronic health records: demonstrating the case for natural language processing},
\newblock \bibinfo{journal}{JCO clinical cancer informatics} \bibinfo{volume}{5} (\bibinfo{year}{2021}) \bibinfo{pages}{1005--1014}.
\bibitem[{Hernandez-Boussard et~al.(2020)Hernandez-Boussard, Blayney, and Brooks}]{hernandez2020leveraging}
\bibinfo{author}{T.~Hernandez-Boussard}, \bibinfo{author}{D.~W. Blayney}, \bibinfo{author}{J.~D. Brooks},
\newblock \bibinfo{title}{Leveraging digital data to inform and improve quality cancer care},
\newblock \bibinfo{journal}{Cancer Epidemiology, Biomarkers \& Prevention} \bibinfo{volume}{29} (\bibinfo{year}{2020}) \bibinfo{pages}{816--822}.
\bibitem[{Bozkurt et~al.(2022)Bozkurt, Magnani, Seneviratne, Brooks, and Hernandez-Boussard}]{bozkurt2022expanding}
\bibinfo{author}{S.~Bozkurt}, \bibinfo{author}{C.~J. Magnani}, \bibinfo{author}{M.~G. Seneviratne}, \bibinfo{author}{J.~D. Brooks}, \bibinfo{author}{T.~Hernandez-Boussard},
\newblock \bibinfo{title}{Expanding the secondary use of prostate cancer real world data: Automated classifiers for clinical and pathological stage},
\newblock \bibinfo{journal}{Frontiers in Digital Health} \bibinfo{volume}{4} (\bibinfo{year}{2022}) \bibinfo{pages}{793316}.
\bibitem[{Bozkurt et~al.(2020)Bozkurt, Paul, Coquet, Sun, Banerjee, Brooks, and Hernandez-Boussard}]{bozkurt2020phenotyping}
\bibinfo{author}{S.~Bozkurt}, \bibinfo{author}{R.~Paul}, \bibinfo{author}{J.~Coquet}, \bibinfo{author}{R.~Sun}, \bibinfo{author}{I.~Banerjee}, \bibinfo{author}{J.~D. Brooks}, \bibinfo{author}{T.~Hernandez-Boussard},
\newblock \bibinfo{title}{Phenotyping severity of patient-centered outcomes using clinical notes: A prostate cancer use case},
\newblock \bibinfo{journal}{Learning Health Systems} \bibinfo{volume}{4} (\bibinfo{year}{2020}) \bibinfo{pages}{e10237}.
\bibitem[{Bozkurt et~al.(2019)Bozkurt, Kan, Ferrari, Rubin, Blayney, Hernandez-Boussard, and Brooks}]{bozkurt2019possible}
\bibinfo{author}{S.~Bozkurt}, \bibinfo{author}{K.~M. Kan}, \bibinfo{author}{M.~K. Ferrari}, \bibinfo{author}{D.~L. Rubin}, \bibinfo{author}{D.~W. Blayney}, \bibinfo{author}{T.~Hernandez-Boussard}, \bibinfo{author}{J.~D. Brooks},
\newblock \bibinfo{title}{Is it possible to automatically assess pretreatment digital rectal examination documentation using natural language processing? a single-centre retrospective study},
\newblock \bibinfo{journal}{BMJ open} \bibinfo{volume}{9} (\bibinfo{year}{2019}) \bibinfo{pages}{e027182}.
\bibitem[{Coquet et~al.(2019)Coquet, Bozkurt, Kan, Ferrari, Blayney, Brooks, and Hernandez-Boussard}]{coquet2019comparison}
\bibinfo{author}{J.~Coquet}, \bibinfo{author}{S.~Bozkurt}, \bibinfo{author}{K.~M. Kan}, \bibinfo{author}{M.~K. Ferrari}, \bibinfo{author}{D.~W. Blayney}, \bibinfo{author}{J.~D. Brooks}, \bibinfo{author}{T.~Hernandez-Boussard},
\newblock \bibinfo{title}{Comparison of orthogonal nlp methods for clinical phenotyping and assessment of bone scan utilization among prostate cancer patients},
\newblock \bibinfo{journal}{Journal of biomedical informatics} \bibinfo{volume}{94} (\bibinfo{year}{2019}) \bibinfo{pages}{103184}.
\bibitem[{Zhu et~al.(2019)Zhu, Lenert, Bunnell, Obeid, Jefferson, and Halbert}]{zhu2019automatically}
\bibinfo{author}{V.~J. Zhu}, \bibinfo{author}{L.~A. Lenert}, \bibinfo{author}{B.~E. Bunnell}, \bibinfo{author}{J.~S. Obeid}, \bibinfo{author}{M.~Jefferson}, \bibinfo{author}{C.~H. Halbert},
\newblock \bibinfo{title}{Automatically identifying social isolation from clinical narratives for patients with prostate cancer},
\newblock \bibinfo{journal}{BMC medical informatics and decision making} \bibinfo{volume}{19} (\bibinfo{year}{2019}) \bibinfo{pages}{1--9}.
\bibitem[{Banerjee et~al.(2019)Banerjee, Li, Seneviratne, Ferrari, Seto, Brooks, Rubin, and Hernandez-Boussard}]{banerjee2019weakly}
\bibinfo{author}{I.~Banerjee}, \bibinfo{author}{K.~Li}, \bibinfo{author}{M.~Seneviratne}, \bibinfo{author}{M.~Ferrari}, \bibinfo{author}{T.~Seto}, \bibinfo{author}{J.~D. Brooks}, \bibinfo{author}{D.~L. Rubin}, \bibinfo{author}{T.~Hernandez-Boussard},
\newblock \bibinfo{title}{Weakly supervised natural language processing for assessing patient-centered outcome following prostate cancer treatment},
\newblock \bibinfo{journal}{JAMIA open} \bibinfo{volume}{2} (\bibinfo{year}{2019}) \bibinfo{pages}{150--159}.
\bibitem[{Huang et~al.(2023)Huang, Lim, Gu, Han, Fang, San~Chia, Bei, Tham, Ho, Yuen et~al.}]{huang2023natural}
\bibinfo{author}{H.~Huang}, \bibinfo{author}{F.~X.~Y. Lim}, \bibinfo{author}{G.~T. Gu}, \bibinfo{author}{M.~J. Han}, \bibinfo{author}{A.~H.~S. Fang}, \bibinfo{author}{E.~H. San~Chia}, \bibinfo{author}{E.~Y.~T. Bei}, \bibinfo{author}{S.~Z. Tham}, \bibinfo{author}{H.~S.~S. Ho}, \bibinfo{author}{J.~S.~P. Yuen}, et~al.,
\newblock \bibinfo{title}{Natural language processing in urology: automated extraction of clinical information from histopathology reports of uro-oncology procedures},
\newblock \bibinfo{journal}{Heliyon} \bibinfo{volume}{9} (\bibinfo{year}{2023}).
\bibitem[{McGowan et~al.(2024)McGowan, Martins, Keen, Whitehead, Davis, Pathiraja, Bolton, and Baldwin}]{mcgowan2024can}
\bibinfo{author}{M.~McGowan}, \bibinfo{author}{F.~C. Martins}, \bibinfo{author}{J.-L. Keen}, \bibinfo{author}{A.~Whitehead}, \bibinfo{author}{E.~Davis}, \bibinfo{author}{P.~Pathiraja}, \bibinfo{author}{H.~Bolton}, \bibinfo{author}{P.~Baldwin},
\newblock \bibinfo{title}{Can natural language processing be effectively applied for audit data analysis in gynaecological oncology at a uk cancer centre?},
\newblock \bibinfo{journal}{International Journal of Medical Informatics} \bibinfo{volume}{182} (\bibinfo{year}{2024}) \bibinfo{pages}{105306}.
\bibitem[{Laios et~al.(2023)Laios, Kalampokis, Mamalis, Tarabanis, Nugent, Thangavelu, Theophilou, and De~Jong}]{laios2023roberta}
\bibinfo{author}{A.~Laios}, \bibinfo{author}{E.~Kalampokis}, \bibinfo{author}{M.~E. Mamalis}, \bibinfo{author}{C.~Tarabanis}, \bibinfo{author}{D.~Nugent}, \bibinfo{author}{A.~Thangavelu}, \bibinfo{author}{G.~Theophilou}, \bibinfo{author}{D.~De~Jong},
\newblock \bibinfo{title}{Roberta-assisted outcome prediction in ovarian cancer cytoreductive surgery using operative notes},
\newblock \bibinfo{journal}{Cancer Control} \bibinfo{volume}{30} (\bibinfo{year}{2023}) \bibinfo{pages}{10732748231209892}.
\bibitem[{Macchia et~al.(2022)Macchia, Ferrandina, Patarnello, Autorino, Masciocchi, Pisapia, Calvani, Iacomini, Cesario, Boldrini et~al.}]{macchia2022multidisciplinary}
\bibinfo{author}{G.~Macchia}, \bibinfo{author}{G.~Ferrandina}, \bibinfo{author}{S.~Patarnello}, \bibinfo{author}{R.~Autorino}, \bibinfo{author}{C.~Masciocchi}, \bibinfo{author}{V.~Pisapia}, \bibinfo{author}{C.~Calvani}, \bibinfo{author}{C.~Iacomini}, \bibinfo{author}{A.~Cesario}, \bibinfo{author}{L.~Boldrini}, et~al.,
\newblock \bibinfo{title}{Multidisciplinary tumor board smart virtual assistant in locally advanced cervical cancer: A proof of concept},
\newblock \bibinfo{journal}{Frontiers in Oncology} \bibinfo{volume}{11} (\bibinfo{year}{2022}) \bibinfo{pages}{797454}.
\bibitem[{Yoo et~al.(2022)Yoo, Yoon, Boo, Kim, Kim, Paeng, Yoo, Choi, Kim, Ryoo et~al.}]{yoo2022transforming}
\bibinfo{author}{S.~Yoo}, \bibinfo{author}{E.~Yoon}, \bibinfo{author}{D.~Boo}, \bibinfo{author}{B.~Kim}, \bibinfo{author}{S.~Kim}, \bibinfo{author}{J.~C. Paeng}, \bibinfo{author}{I.~R. Yoo}, \bibinfo{author}{I.~Y. Choi}, \bibinfo{author}{K.~Kim}, \bibinfo{author}{H.~G. Ryoo}, et~al.,
\newblock \bibinfo{title}{Transforming thyroid cancer diagnosis and staging information from unstructured reports to the observational medical outcome partnership common data model},
\newblock \bibinfo{journal}{Applied Clinical Informatics} \bibinfo{volume}{13} (\bibinfo{year}{2022}) \bibinfo{pages}{521--531}.
\bibitem[{Pathak et~al.(2023)Pathak, Yu, Paredes, Monsour, Rocha, Brito, Ospina, and Wu}]{pathak2023extracting}
\bibinfo{author}{A.~Pathak}, \bibinfo{author}{Z.~Yu}, \bibinfo{author}{D.~Paredes}, \bibinfo{author}{E.~P. Monsour}, \bibinfo{author}{A.~O. Rocha}, \bibinfo{author}{J.~P. Brito}, \bibinfo{author}{N.~S. Ospina}, \bibinfo{author}{Y.~Wu},
\newblock \bibinfo{title}{Extracting thyroid nodules characteristics from ultrasound reports using transformer-based natural language processing methods},
\newblock in: \bibinfo{booktitle}{AMIA Annual Symposium Proceedings}, volume \bibinfo{volume}{2023}, \bibinfo{organization}{American Medical Informatics Association}, \bibinfo{year}{2023}, p. \bibinfo{pages}{1193}.
\bibitem[{Ali et~al.(2022)Ali, Strafford, Dobbs, Fonferko-Shadrach, Lacey, Pickrell, Hutchings, and Whitaker}]{ali2022development}
\bibinfo{author}{S.~R. Ali}, \bibinfo{author}{H.~Strafford}, \bibinfo{author}{T.~D. Dobbs}, \bibinfo{author}{B.~Fonferko-Shadrach}, \bibinfo{author}{A.~S. Lacey}, \bibinfo{author}{W.~O. Pickrell}, \bibinfo{author}{H.~A. Hutchings}, \bibinfo{author}{I.~S. Whitaker},
\newblock \bibinfo{title}{Development and validation of an automated basal cell carcinoma histopathology information extraction system using natural language processing},
\newblock \bibinfo{journal}{Frontiers in Surgery} \bibinfo{volume}{9} (\bibinfo{year}{2022}) \bibinfo{pages}{870494}.
\bibitem[{Malke et~al.(2019)Malke, Jin, Camp, Lari, Kell, Simon, Prieto, Gershenwald, and Haydu}]{malke2019enhancing}
\bibinfo{author}{J.~C. Malke}, \bibinfo{author}{S.~Jin}, \bibinfo{author}{S.~P. Camp}, \bibinfo{author}{B.~Lari}, \bibinfo{author}{T.~Kell}, \bibinfo{author}{J.~M. Simon}, \bibinfo{author}{V.~G. Prieto}, \bibinfo{author}{J.~E. Gershenwald}, \bibinfo{author}{L.~E. Haydu},
\newblock \bibinfo{title}{Enhancing case capture, quality, and completeness of primary melanoma pathology records via natural language processing},
\newblock \bibinfo{journal}{JCO clinical cancer informatics} \bibinfo{volume}{3} (\bibinfo{year}{2019}) \bibinfo{pages}{1--11}.
\bibitem[{Senders et~al.(2019)Senders, Karhade, Cote, Mehrtash, Lamba, DiRisio, Muskens, Gormley, Smith, Broekman et~al.}]{senders2019natural}
\bibinfo{author}{J.~T. Senders}, \bibinfo{author}{A.~V. Karhade}, \bibinfo{author}{D.~J. Cote}, \bibinfo{author}{A.~Mehrtash}, \bibinfo{author}{N.~Lamba}, \bibinfo{author}{A.~DiRisio}, \bibinfo{author}{I.~S. Muskens}, \bibinfo{author}{W.~B. Gormley}, \bibinfo{author}{T.~R. Smith}, \bibinfo{author}{M.~L. Broekman}, et~al.,
\newblock \bibinfo{title}{Natural language processing for automated quantification of brain metastases reported in free-text radiology reports},
\newblock \bibinfo{journal}{JCO clinical cancer informatics} \bibinfo{volume}{3} (\bibinfo{year}{2019}) \bibinfo{pages}{1--9}.
\bibitem[{Lindvall et~al.(2022)Lindvall, Deng, Moseley, Agaronnik, El-Jawahri, Paasche-Orlow, Lakin, Volandes, Tulsky, Investigators et~al.}]{lindvall2022natural}
\bibinfo{author}{C.~Lindvall}, \bibinfo{author}{C.-Y. Deng}, \bibinfo{author}{E.~Moseley}, \bibinfo{author}{N.~Agaronnik}, \bibinfo{author}{A.~El-Jawahri}, \bibinfo{author}{M.~K. Paasche-Orlow}, \bibinfo{author}{J.~R. Lakin}, \bibinfo{author}{A.~Volandes}, \bibinfo{author}{J.~A. Tulsky}, \bibinfo{author}{A.-P. Investigators}, et~al.,
\newblock \bibinfo{title}{Natural language processing to identify advance care planning documentation in a multisite pragmatic clinical trial},
\newblock \bibinfo{journal}{Journal of pain and symptom management} \bibinfo{volume}{63} (\bibinfo{year}{2022}) \bibinfo{pages}{e29--e36}.
\bibitem[{Lindvall et~al.(2019)Lindvall, Lilley, Zupanc, Chien, Udelsman, Walling, Cooper, and Tulsky}]{lindvall2019natural}
\bibinfo{author}{C.~Lindvall}, \bibinfo{author}{E.~J. Lilley}, \bibinfo{author}{S.~N. Zupanc}, \bibinfo{author}{I.~Chien}, \bibinfo{author}{B.~V. Udelsman}, \bibinfo{author}{A.~Walling}, \bibinfo{author}{Z.~Cooper}, \bibinfo{author}{J.~A. Tulsky},
\newblock \bibinfo{title}{Natural language processing to assess end-of-life quality indicators in cancer patients receiving palliative surgery},
\newblock \bibinfo{journal}{Journal of palliative medicine} \bibinfo{volume}{22} (\bibinfo{year}{2019}) \bibinfo{pages}{183--187}.
\bibitem[{Ernecoff et~al.(2019)Ernecoff, Wessell, Hanson, Lee, Shea, Dusetzina, Weinberger, and Bennett}]{ernecoff2019electronic}
\bibinfo{author}{N.~C. Ernecoff}, \bibinfo{author}{K.~L. Wessell}, \bibinfo{author}{L.~C. Hanson}, \bibinfo{author}{A.~M. Lee}, \bibinfo{author}{C.~M. Shea}, \bibinfo{author}{S.~B. Dusetzina}, \bibinfo{author}{M.~Weinberger}, \bibinfo{author}{A.~V. Bennett},
\newblock \bibinfo{title}{Electronic health record phenotypes for identifying patients with late-stage disease: A method for research and clinical application},
\newblock \bibinfo{journal}{Journal of general internal medicine} \bibinfo{volume}{34} (\bibinfo{year}{2019}) \bibinfo{pages}{2818--2823}.
\bibitem[{DiMartino et~al.(2022)DiMartino, Miano, Wessell, Bohac, and Hanson}]{dimartino2022identification}
\bibinfo{author}{L.~DiMartino}, \bibinfo{author}{T.~Miano}, \bibinfo{author}{K.~Wessell}, \bibinfo{author}{B.~Bohac}, \bibinfo{author}{L.~C. Hanson},
\newblock \bibinfo{title}{Identification of uncontrolled symptoms in cancer patients using natural language processing},
\newblock \bibinfo{journal}{Journal of pain and symptom management} \bibinfo{volume}{63} (\bibinfo{year}{2022}) \bibinfo{pages}{610--617}.
\bibitem[{Laurent et~al.(2023)Laurent, Craynest, Thobois, and Hajjaji}]{laurent2023automatic}
\bibinfo{author}{G.~Laurent}, \bibinfo{author}{F.~Craynest}, \bibinfo{author}{M.~Thobois}, \bibinfo{author}{N.~Hajjaji},
\newblock \bibinfo{title}{Automatic classification of tumor response from radiology reports with rule-based natural language processing integrated into the clinical oncology workflow},
\newblock \bibinfo{journal}{JCO Clinical Cancer Informatics} \bibinfo{volume}{7} (\bibinfo{year}{2023}) \bibinfo{pages}{e2200139}.
\bibitem[{Lin et~al.(2023)Lin, Zwolinski, Wu, La, Goryachev, Huhmann, Yildrim, Tuck, Elbers, Brophy et~al.}]{lin2023machine}
\bibinfo{author}{E.~Lin}, \bibinfo{author}{R.~Zwolinski}, \bibinfo{author}{J.~T.-Y. Wu}, \bibinfo{author}{J.~La}, \bibinfo{author}{S.~Goryachev}, \bibinfo{author}{L.~Huhmann}, \bibinfo{author}{C.~Yildrim}, \bibinfo{author}{D.~P. Tuck}, \bibinfo{author}{D.~C. Elbers}, \bibinfo{author}{M.~T. Brophy}, et~al.,
\newblock \bibinfo{title}{Machine learning-based natural language processing to extract pd-l1 expression levels from clinical notes},
\newblock \bibinfo{journal}{Health Informatics Journal} \bibinfo{volume}{29} (\bibinfo{year}{2023}) \bibinfo{pages}{14604582231198021}.
\bibitem[{Naseri et~al.(2021)Naseri, Kafi, Skamene, Tolba, Faye, Ramia, Khriguian, and Kildea}]{naseri2021development}
\bibinfo{author}{H.~Naseri}, \bibinfo{author}{K.~Kafi}, \bibinfo{author}{S.~Skamene}, \bibinfo{author}{M.~Tolba}, \bibinfo{author}{M.~D. Faye}, \bibinfo{author}{P.~Ramia}, \bibinfo{author}{J.~Khriguian}, \bibinfo{author}{J.~Kildea},
\newblock \bibinfo{title}{Development of a generalizable natural language processing pipeline to extract physician-reported pain from clinical reports: Generated using publicly-available datasets and tested on institutional clinical reports for cancer patients with bone metastases},
\newblock \bibinfo{journal}{Journal of Biomedical Informatics} \bibinfo{volume}{120} (\bibinfo{year}{2021}) \bibinfo{pages}{103864}.
\bibitem[{Ahmad et~al.(2023)Ahmad, Liu, Khan, Jiang, and Burhan}]{ahmad2023bir}
\bibinfo{author}{P.~N. Ahmad}, \bibinfo{author}{Y.~Liu}, \bibinfo{author}{K.~Khan}, \bibinfo{author}{T.~Jiang}, \bibinfo{author}{U.~Burhan},
\newblock \bibinfo{title}{Bir: Biomedical information retrieval system for cancer treatment in electronic health record using transformers},
\newblock \bibinfo{journal}{Sensors} \bibinfo{volume}{23} (\bibinfo{year}{2023}) \bibinfo{pages}{9355}.
\bibitem[{De~Angeli et~al.(2022)De~Angeli, Gao, Danciu, Durbin, Wu, Stroup, Doherty, Schwartz, Wiggins, Damesyn et~al.}]{de2022class}
\bibinfo{author}{K.~De~Angeli}, \bibinfo{author}{S.~Gao}, \bibinfo{author}{I.~Danciu}, \bibinfo{author}{E.~B. Durbin}, \bibinfo{author}{X.-C. Wu}, \bibinfo{author}{A.~Stroup}, \bibinfo{author}{J.~Doherty}, \bibinfo{author}{S.~Schwartz}, \bibinfo{author}{C.~Wiggins}, \bibinfo{author}{M.~Damesyn}, et~al.,
\newblock \bibinfo{title}{Class imbalance in out-of-distribution datasets: Improving the robustness of the textcnn for the classification of rare cancer types},
\newblock \bibinfo{journal}{Journal of biomedical informatics} \bibinfo{volume}{125} (\bibinfo{year}{2022}) \bibinfo{pages}{103957}.
\bibitem[{Cohen et~al.(2023)Cohen, Rosic, Harrison, Richey, Nemeth, Ambwani, Miksad, Haaland, and Jiang}]{cohen2023natural}
\bibinfo{author}{A.~B. Cohen}, \bibinfo{author}{A.~Rosic}, \bibinfo{author}{K.~Harrison}, \bibinfo{author}{M.~Richey}, \bibinfo{author}{S.~Nemeth}, \bibinfo{author}{G.~Ambwani}, \bibinfo{author}{R.~Miksad}, \bibinfo{author}{B.~Haaland}, \bibinfo{author}{C.~Jiang},
\newblock \bibinfo{title}{A natural language processing algorithm to improve completeness of ecog performance status in real-world data},
\newblock \bibinfo{journal}{Applied Sciences} \bibinfo{volume}{13} (\bibinfo{year}{2023}) \bibinfo{pages}{6209}.
\bibitem[{Koleck et~al.(2021)Koleck, Topaz, Tatonetti, George, Miaskowski, Smaldone, and Bakken}]{koleck2021characterizing}
\bibinfo{author}{T.~A. Koleck}, \bibinfo{author}{M.~Topaz}, \bibinfo{author}{N.~P. Tatonetti}, \bibinfo{author}{M.~George}, \bibinfo{author}{C.~Miaskowski}, \bibinfo{author}{A.~Smaldone}, \bibinfo{author}{S.~Bakken},
\newblock \bibinfo{title}{Characterizing shared and distinct symptom clusters in common chronic conditions through natural language processing of nursing notes},
\newblock \bibinfo{journal}{Research in nursing \& health} \bibinfo{volume}{44} (\bibinfo{year}{2021}) \bibinfo{pages}{906--919}.
\bibitem[{Guan et~al.(2019)Guan, Cho, Petro, Zhang, Pasche, and Topaloglu}]{guan2019natural}
\bibinfo{author}{M.~Guan}, \bibinfo{author}{S.~Cho}, \bibinfo{author}{R.~Petro}, \bibinfo{author}{W.~Zhang}, \bibinfo{author}{B.~Pasche}, \bibinfo{author}{U.~Topaloglu},
\newblock \bibinfo{title}{Natural language processing and recurrent network models for identifying genomic mutation-associated cancer treatment change from patient progress notes},
\newblock \bibinfo{journal}{JAMIA open} \bibinfo{volume}{2} (\bibinfo{year}{2019}) \bibinfo{pages}{139--149}.
\bibitem[{Mashima et~al.(2022)Mashima, Tamura, Kunikata, Tada, Yamada, Tanigawa, Hayakawa, Tanabe, and Yokoi}]{mashima2022using}
\bibinfo{author}{Y.~Mashima}, \bibinfo{author}{T.~Tamura}, \bibinfo{author}{J.~Kunikata}, \bibinfo{author}{S.~Tada}, \bibinfo{author}{A.~Yamada}, \bibinfo{author}{M.~Tanigawa}, \bibinfo{author}{A.~Hayakawa}, \bibinfo{author}{H.~Tanabe}, \bibinfo{author}{H.~Yokoi},
\newblock \bibinfo{title}{Using natural language processing techniques to detect adverse events from progress notes due to chemotherapy},
\newblock \bibinfo{journal}{Cancer Informatics} \bibinfo{volume}{21} (\bibinfo{year}{2022}) \bibinfo{pages}{11769351221085064}.
\bibitem[{Li et~al.(2022)Li, da~Costa~Jr, Guffey, Milner, Allam, Kurian, Novoa, Poche, Bandyo, Granada et~al.}]{li2022developing}
\bibinfo{author}{A.~Li}, \bibinfo{author}{W.~L. da~Costa~Jr}, \bibinfo{author}{D.~Guffey}, \bibinfo{author}{E.~M. Milner}, \bibinfo{author}{A.~K. Allam}, \bibinfo{author}{K.~M. Kurian}, \bibinfo{author}{F.~J. Novoa}, \bibinfo{author}{M.~D. Poche}, \bibinfo{author}{R.~Bandyo}, \bibinfo{author}{C.~Granada}, et~al.,
\newblock \bibinfo{title}{Developing and optimizing a computable phenotype for incident venous thromboembolism in a longitudinal cohort of patients with cancer},
\newblock \bibinfo{journal}{Research and practice in thrombosis and haemostasis} \bibinfo{volume}{6} (\bibinfo{year}{2022}) \bibinfo{pages}{e12733}.
\bibitem[{Hong et~al.(2020)Hong, Fairchild, Tanksley, Palta, and Tenenbaum}]{hong2020natural}
\bibinfo{author}{J.~C. Hong}, \bibinfo{author}{A.~T. Fairchild}, \bibinfo{author}{J.~P. Tanksley}, \bibinfo{author}{M.~Palta}, \bibinfo{author}{J.~D. Tenenbaum},
\newblock \bibinfo{title}{Natural language processing for abstraction of cancer treatment toxicities: accuracy versus human experts},
\newblock \bibinfo{journal}{JAMIA open} \bibinfo{volume}{3} (\bibinfo{year}{2020}) \bibinfo{pages}{513--517}.
\bibitem[{Mu{\~n}oz et~al.(2023)Mu{\~n}oz, Souto, Lecumberri, Obispo, Sanchez, Aparicio, Aguayo, Gutierrez, Palomo, Fanjul et~al.}]{munoz2023development}
\bibinfo{author}{A.~J. Mu{\~n}oz}, \bibinfo{author}{J.~C. Souto}, \bibinfo{author}{R.~Lecumberri}, \bibinfo{author}{B.~Obispo}, \bibinfo{author}{A.~Sanchez}, \bibinfo{author}{J.~Aparicio}, \bibinfo{author}{C.~Aguayo}, \bibinfo{author}{D.~Gutierrez}, \bibinfo{author}{A.~G. Palomo}, \bibinfo{author}{V.~Fanjul}, et~al.,
\newblock \bibinfo{title}{Development of a predictive model of venous thromboembolism recurrence in anticoagulated cancer patients using machine learning},
\newblock \bibinfo{journal}{Thrombosis Research} \bibinfo{volume}{228} (\bibinfo{year}{2023}) \bibinfo{pages}{181--188}.
\bibitem[{Zitu et~al.(2023)Zitu, Zhang, Owen, Chiang, and Li}]{zitu2023generalizability}
\bibinfo{author}{M.~M. Zitu}, \bibinfo{author}{S.~Zhang}, \bibinfo{author}{D.~H. Owen}, \bibinfo{author}{C.~Chiang}, \bibinfo{author}{L.~Li},
\newblock \bibinfo{title}{Generalizability of machine learning methods in detecting adverse drug events from clinical narratives in electronic medical records},
\newblock \bibinfo{journal}{Frontiers in Pharmacology} \bibinfo{volume}{14} (\bibinfo{year}{2023}) \bibinfo{pages}{1218679}.
\bibitem[{Iannantuono et~al.(2023)Iannantuono, Bracken-Clarke, Floudas, Roselli, Gulley, and Karzai}]{iannantuono2023applications}
\bibinfo{author}{G.~M. Iannantuono}, \bibinfo{author}{D.~Bracken-Clarke}, \bibinfo{author}{C.~S. Floudas}, \bibinfo{author}{M.~Roselli}, \bibinfo{author}{J.~L. Gulley}, \bibinfo{author}{F.~Karzai},
\newblock \bibinfo{title}{Applications of large language models in cancer care: current evidence and future perspectives},
\newblock \bibinfo{journal}{Frontiers in Oncology} \bibinfo{volume}{13} (\bibinfo{year}{2023}) \bibinfo{pages}{1268915}.

\end{thebibliography}

\end{document}